%% file: main.tex
\documentclass[review]{elsarticle}

\usepackage[hyphens]{url}
\usepackage{lineno,hyperref}
\usepackage{amsmath}
\usepackage[utf8]{inputenc} 
\usepackage[T1]{fontenc}    
\usepackage{booktabs}       
\usepackage{amsfonts}       
\usepackage{nicefrac}       
\usepackage{microtype}      
\usepackage{float}
\usepackage{algorithm}
\usepackage[noend]{algpseudocode}
\usepackage{bm}
\usepackage{mathtools}
\usepackage{multicol}
\usepackage{todonotes}
\usepackage{graphicx}
\usepackage{footnote}
\usepackage{common}

\usepackage{lmodern}
\usepackage{tikz}
\usepackage{glossaries}
\usepackage[scientific-notation=true]{siunitx}
\usepackage{subcaption}

\usetikzlibrary{calc}
\usetikzlibrary{positioning}
\usetikzlibrary{chains,fit,shapes}

\modulolinenumbers[5]

\newacronym{rnn}{RNN}{Recurrent Neural Network}
\newacronym{cwrnn}{CW-RNN}{Clockwork Recurrent Neural Network}
\newacronym{lmn}{LMN}{Linear Memory Network}
\newacronym[firstplural=Long-Short Term Memory (LSTMs)]{lstm}{LSTM}{Long-Short Term Memory}
\newacronym{gru}{GRU}{Gated Recurrent Unit}
\newacronym{bptt}{BPTT}{Backpropagation Through Time}
\newacronym{cwlmn}{CW-LMN}{Clockwork Linear Memory Network}

\journal{Neural Networks}

\begin{document}

    \begin{frontmatter}
        \title{Encoding-based Memory Modules for Recurrent Neural Networks}

        \author[1]{Antonio Carta\corref{cor1}}
        \cortext[cor1]{corresponding author}
        \ead{antonio.carta@di.unipi.it}
        \author[2]{Alessandro Sperduti}
        \ead{sperduti@math.unipd.it}
        \author[1]{Davide Bacciu}
        \ead{bacciu@di.unipi.it}
        
        \address[1]{Department of Computer Science, University of Pisa}
        \address[2]{Department of Mathematics, University of Padova}


        \begin{abstract}
        Learning to solve sequential tasks with recurrent models requires the ability to memorize long sequences and to extract task-relevant features from them. In this paper, we study the memorization subtask from the point of view of the design and training of recurrent neural networks. We propose a new model, the Linear Memory Network, which features an encoding-based memorization component built with a linear autoencoder for sequences. We extend the memorization component with a modular memory that encodes the hidden state sequence at different sampling frequencies. Additionally, we provide a specialized training algorithm that initializes the memory to efficiently encode the hidden activations of the network. The experimental results on synthetic and real-world datasets show that specializing the training algorithm to train the memorization component always improves the final performance whenever the memorization of long sequences is necessary to solve the problem.
        \end{abstract}


    \end{frontmatter}


    \input{body.tex}

    \bibliography{biblio}
    
    \appendix
    \input{appendix.tex}

\end{document}

%% file: body.tex
\section{Introduction} \label{sec:intro}
Sequential data, such as music, speech, text, videos, are ubiquitous and present several challenges. In recent times. recurrent neural networks have seen a widespread use in sequential domains such as natural language processing~\cite{nlp_review17}, speech recognition~\cite{speech_review13} or time series analysis~\cite{time_series_review14}.
Processing sequential data with recurrent models requires the ability to solve two different subproblems: the problem of the extraction of informative features from the input and the ability to memorize the relevant information efficiently. Therefore, to be effective a recurrent model must be expressive enough to solve both of these problems while being easy to train.
To reduce the complexity of recurrent models without hindering their expressivity, we propose a conceptual separation of sequential problems into two subtasks, which are generic problems that must be accounted in each sequential problem solved with recurrent models:
\begin{description}
    \item[functional subtask], that is the mapping between the subsequence up to a certain timestep and a set of task-relevant features.
    \item[memorization subtask], that is the mapping responsible for the update of the internal state of the model, and therefore the memorization of the previously computed features.
\end{description}
\begin{figure}
    \centering
    \input{img/fun_mem_arch.tex}
    \caption{High-level overview of the functional-memorization separation for recurrent neural networks.}
    \label{fig:fun_mem_separation}
\end{figure}
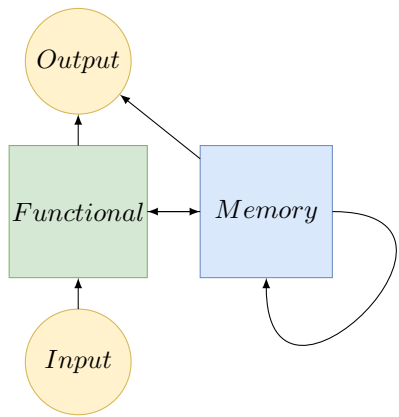

A high-level overview of the separation is shown in Figure \ref{fig:fun_mem_separation}. Notice that under this separation only the memorization component is recurrent. This conceptual separation can guide the design of recurrent architectures by simplifying their architecture and by investigating their ability to solve each subproblem separately, as we will show in Section \ref{sec:lmn} and \ref{sec:lmn_train}.

In this paper, we show how to build a recurrent neural network, the Linear Memory Network (LMN), with separate memorization and functional components. To address the memorization subtask, which is a difficult task to solve due to the recurrent nature of the problem, we simplify the architecture of the memorization component by using a linear autoencoder. Building on the architectural separation, we propose a novel training algorithm that attempts to solve the two subproblems separately. The functional-memory separation can be exploited to define more complex memorization mechanisms, such as the MultiScale LMN (MS-LMN), that extends the memorization component with a modular memory, better suited to model long-range dependencies. Correspondingly, we introduce a specialized training algorithm for the MS-LMN that works incrementally by alternating training of the functional and of the memorization component, and that solves some of the limitations of the LMN memory training. The experimental results on several synthetic and real-world datasets show that the models achieve a competitive performance whenever the memorization of long sequences is fundamental to solve the tasks. More importantly, the proposed training algorithm always improves the performance compared to traditional end-to-end backpropagation.

This paper represents an extension of a previous work~\cite{lmn_carta_icann19} presented as a conference paper. The original contributions of this paper are the introduction of the MultiScale LMN as an alternative memorization module, together with its corresponding specialized training algorithm and experimental results, and an in-depth presentation of the proposed framework based on the separation between the memory and functional components.

The paper is organized as follows: Section \ref{sec:background} introduces background material, including the notation, a short review of RNN architectures and their relevance compared to the proposed approach, and the linear autoencoder for sequences. Section \ref{sec:lmn} describes the LMN and MS-LMN architectures. Section \ref{sec:lmn_train} presents the specialized training algorithm used to train the memorization component of the LMN and MS-LMN. Section \ref{sec:experiments} shows the experimental results on several synthetic and real-world datasets. Finally, we draw the conclusions in Section \ref{sec:conclusion} and highlight possible avenues for future work.

\section{Background} \label{sec:background}
\subsection{Notation}\label{sec:notation}
We represent sequences as an ordered list of vectors $\vx^1, \hdots, \vx^l$, where $l$ is the number of elements in the sequence.
To simplify the notation, biases are omitted from the equations. We refer the reader to the following table for the notation used throughout the paper.

\begin{tabular}{ll}
    \toprule
    Symbol                   & Description  \\
    \midrule
    $\vx$                    & vector (bold and lower-case letters)  \\
    $\mW$                    & matrix (bold and upper-case letters)  \\
    $\mW^{ab}$               & parameters of the layer with input $\va$ and output $\vb$ \\
    $\vx^t$                  & $t^{th}$ element of a sequence  \\
    $\mW^\top$               & transpose of $\mW$   \\
    $ N_x$                   & size of vector $\vx$ \\
    $\tilde{\vx}$            & approximate reconstruction of $\vx$ \\
    $\mW[start:end]$         & submatrix of $\mW$ with rows $start, \hdots, end$ \\
    $\mW[:end]$              & submatrix of $\mW$ with rows $1, \hdots, end$ \\
    $\mW[start:]$            & submatrix of $\mW$ with rows $start, \hdots, n_{rows}$ \\
    \bottomrule
\end{tabular}

\subsection{Recurrent Neural Networks}
While to our knowledge this paper is the first that defines an explicit separation between the functional and memorization component, most of the recurrent neural models in the literature try to address the memorization problem with different solutions.

For example, the \emph{Vanilla RNN}~\cite{elman_rnn} solves the memorization and functional subtasks concurrently, using a single hidden state that is updated with a nonlinear equation. The result is a highly expressive model, even Turing equivalent \cite{siegelmann1995computational}. Unfortunately, RNNs are difficult to train due to the vanishing gradient problem \cite{hochreiter1998vanishing}.

\emph{Gated architectures} like LSTM\cite{hochreiter_lstm97,Cummins2000,lstm_odyssey_greff17} and GRU\cite{gru_Chung2014} partially solve the vanishing gradient problem by updating an internal cell state using the sum operation. A set of gates are used to control the cell state and can be used to block updates to the cell or to forget previous activations by resetting the corresponding hidden units. However, gated architectures are still ineffective for problems that require the memorization of long sequences~\cite{unitary_Arjovsky2015}.

\emph{Hierarchical RNNs} instead model long-term dependencies by allowing to skip hidden state updates, which also mitigates the vanishing gradient problem. This is often achieved either through subsampling~\cite{clockwork_koutnik14,Roberts2018}, learning to skip updates~\cite{hmrnn_Chung2017,Campos2017a} or by introducing skip connections~\cite{Zilly2016}.

Another class of recurrent models called \emph{Memory Augmented Neural Networks (MANN)} defines complex memorization and addressing mechanisms based on stacks \cite{stack_rnn_joulin15,grefensteette_unbuonded_nips15,stack_rnn_nlp_yogatama18} or associative memories \cite{ntm_Graves2014,dnc_nature_Graves2016a,neural_gpu_Kaiser2015}. These models attempt to include some form of explicit memory and have shown promising results on some reasoning and algorithmic tasks. However, complex memorization mechanisms make these models extremely difficult to train using end-to-end backpropagation and highly susceptible to hyperparameters.

Compared to our approach, all these models try to solve the memorization subtask indirectly. However, this is difficult to do in practice, due to the need to learn long-term dependencies and the vanishing gradient problem. Instead, by completely separating the model into two components we can train the memorization separately, as we will show in Section \ref{sec:lmn_train}. To the best of our knowledge, this is the first work that tries to address the memorization subtasks directly on an architectural and training level. This represents an advantage compared to other recurrent architectures since we can design specialized solutions for the memorization problems without hindering the expressivity or trainability of the model. For example, we can simplify the memorization component by removing the nonlinearity in the recurrence without limiting the expressivity of the model, as we will show in Section \ref{sec:lmn}. Other models, like orthogonal RNNs~\cite{unitary_Arjovsky2015,wisdom_full_urnn_nips2016} or IDRNN~\cite{irnn_hinton15} make long-term dependencies easier to learn at the expense of expressivity~\cite{Henaff2016,vorontsov_ortho_rnn_icml17}. By contrast, MANN architectures define complex memorization mechanisms that make them highly expressive but extremely difficult to train, while the separate memory modules of the LMN can be easily trained using the specialized training algorithm proposed in Section \ref{sec:lmn_train}.

\subsection{Linear Autoencoder for Sequences} \label{sec:laes}

The linear autoencoder for sequences (LAES) \cite{sperduti2013linear} is a recurrent linear model that is able to memorize an input sequence by encoding it into a hidden memory state vector recursively updated which represents the entire sequence. Given the memory state, the original sequence can be reconstructed.

Given a sequence $s = \vx^1, \hdots, \vx^l$, where $\vx^i \in \mathbb{R}^a$, a linear autoencoder computes the memory state vector $\vm^t \in \mathbb{R}^p$, i.e. the encoding of the input sequence up to time $t$, using the following equations:

\begin{align}
    \vm^t = \mA\vx^t + \mB\vm^{t-1}  \label{eq:ae_enc} \\
    \begin{bmatrix} \vx^t \\ \vm^{t-1} \end{bmatrix} = \mC\vm^t, \label{eq:ae_dec}
\end{align}

where $p$ is the memory state size, $\mA \in \mathbb{R}^{p \times a}$, $\mB \in \mathbb{R}^{p \times p}$ and $\mC \in \mathbb{R}^{(a+p) \times p}$ are the model parameters. Equation \ref{eq:ae_enc} describes the encoding operation, while Equation \ref{eq:ae_dec} describes the decoding operation.

\subsubsection{Training Algorithm}
The linearity of the LAES allow us to derive the optimal solution with a closed-form equation, as shown in \cite{laes_icann_sperduti06,laes_ecml_sperduti07}. For simplicity, let us assume that the training set consists of a single sequence $\{\vx_1, \hdots, \vx_l\}$ and define $\mM \in \mathbb{R}^{l \times p}$ as the matrix obtained by stacking by rows the memory state vectors of the LAES at each timestep. From Eq. (\ref{eq:ae_enc}) it follows that:

\begin{equation}
\underbrace{
\begin{bmatrix}
{\vm^1}^\top \\ {\vm^2}^\top \\ {\vm^3}^\top \\ \vdots \\ {\vm^l}^\top
\end{bmatrix}}_{\mM} =
\underbrace{\begin{bmatrix}
{\vx^1}^\top & 0 & \hdots & 0 \\
{\vx^2}^\top & {\vx^1}^\top & \hdots & 0 \\
\vdots & \vdots & \ddots & \vdots \\
{\vx^l}^\top & {\vx^{l-1}}^\top & \hdots & {\vx^1}^\top
\end{bmatrix}}_{\mXi}
\underbrace{\begin{bmatrix}
\mA^\top \\ \mA^\top \mB^\top \\ \vdots \\ \mA^\top {\mB^{l-1}}^\top
\end{bmatrix}}_{\mOmega}.
\end{equation}

The matrix $\mXi \in \mathbb{R}^{l \times la}$ contains the reversed subsequences of $s$, while $\mOmega \in \mathbb{R}^{la \times p}$ contains the matrices to encode the input elements for up to $l$ timesteps.
The encoder parameters $\mA$ and $\mB$ can be identified by exploiting the singular value decomposition (SVD) $\mXi = \mV \mSigma \mU^\top$, where imposing $\mU^\top \mOmega = \mI$ yields $\mOmega = \mU$. Given this additional constraint, we can then exploit the structure of $\mXi$ to recover $\mA$, $\mB$, and the decoder parameters $\mC = \begin{bmatrix} \mA^\top \\ \mB^\top \end{bmatrix}$. Specifically, $\mOmega=\mU$ is satisfied by using the matrices
\begin{equation*}
 \mP  \equiv  \left[\begin{array}{c} \mI_{a} \\ \mzero_{a(l - 1)\times a}\end{array}\right],\ \ \mbox{and}\ \ \mR  \equiv  \left[\begin{array}{ll} \mzero_{a\times a(l - 1)} &  \mzero_{a\times a}\\ \mI_{a(l - 1)} & \mzero_{a(l - 1)\times a}\end{array}\right]
\end{equation*}
to define
$\mA\equiv \mU^\top \mP$ and \mbox{$\mB\equiv \mU^\top  \mR \mU$}, where $\mI_{u}$ is the identity matrix of size $u$, and $0_{u\times v}$ is the zero matrix of size $u\times v$.
The algorithm can be easily generalized to multiple sequences by stacking the data matrix $\mXi_q$ for each sequence $s^q$ and padding with zeros to match sequences length, as shown in \cite{sperduti2013linear}.

The optimal solution reported above allows encoding the entire sequence without errors with a minimal number of memory units $p = rank(\mXi)$. Since we fix the number of memory units before computing the LAES, we approximate the optimal solution using the truncated SVD, which introduces some errors during the decoding process.
The computational cost of the training algorithm is dominated by the cost of the truncated SVD decomposition, which for a matrix of size $n \times m$ is $\mathcal{O}(n^2 m)$. Given a dataset of sequences with lengths $l^1, \hdots, l^{S}$, with $l=\sum_{i=1}^S l^i$, $l_{max} = \max_i \{l^i\}$, we have $\mXi \in \mathbb{R}^{l \times l_{max}a}$, which requires $\mathcal{O}(l_{max} l a)$ memory for a dense representation. The memory usage can be reduced for sparse inputs, such as music in a piano roll representation, by using a sparse representation. To mitigate this problem in a more general setting, we approximate the SVD decomposition using the approach proposed in \cite{preatrain_la_Pasa2014}, which computes the SVD decomposition with an iterative algorithm that decomposes $\mXi$ in slices of size $l \times a$ and requires $\mathcal{O}(l a)$ memory.

\section{Linear Memory Network} \label{sec:lmn}
As discussed in Section \ref{sec:intro}, to solve a sequential task we need to solve a memorization and a functional subtask. To impose this separation at an architectural level, we define two separate components. Formally, the functional component $F$ and the memorization component $M$ are two functions defined as follows:
\begin{align*}
    F: Input \times Memory \mapsto Hidden \\
    M: Hidden \times Memory \mapsto Memory,
\end{align*}
where $Input$, $Hidden$, and $Memory$ represent the input, hidden, and memory space respectively. Notice that under this setting, given a fixed memorization component, the functional component can be implemented by a feedforward neural network. Therefore, the recurrence is confined inside the memorization component.

The memorization component $M$ can be implemented as an autoencoder which encodes the entire sequence of activations computed by the functional component $F$. This approach is general and expressive since it allows $F$ to use the encoding of $M$ to reconstruct the entire sequence to compute the new features.

Notice that in principle training $M$ to encode the entire sequence is a difficult problem that always contains long-term dependencies. After the encoder of $M$ computes the final memory state, the decoder must reconstruct the entire sequence. The last state reconstructed is the first state given as input to the encoder. This computation creates a dependency between the first input and the last output, with a length twice the number of timesteps of the input sequence. This makes training $M$ using gradient descent difficult due to the vanishing gradient problem \cite{hochreiter1998vanishing}.

Fortunately, we can sidestep this problem using the LAES solution given in Section \ref{sec:laes}. The LAES can be used in a recurrent neural network as an efficient \emph{encoding-based memorization} module.
To complete our model, we need to combine the LAES with a functional component. In the following sections, we propose a novel recurrent neural network architecture based on the combination of the LAES, as a memorization component, with a feedforward neural network, used as a functional component. We can train the feedforward component of the model to perform the feature extraction while the LAES is responsible only for the memorization of the extracted features.

\subsection{Model Architecture with LAES Memory}

The \emph{Linear Memory Network} (LMN) is a recurrent neural network based on the conceptual separation of sequential problems into a functional and a memorization subtasks. The LMN comprises two separate modules that communicate with each other: a feedforward neural network that implements the functional component, and a linear recurrence, equivalent to a LAES, which implements the memorization component. The model computes a hidden state $\vh^t$ and a memory state $\vm^t$ as follows:
\LMNeq
\begin{figure}
    \centering
    \input{img/lmn.tex}
    \caption{Architecture of the Linear Memory Network, where $\vx^t, \vh^t, \vm^t, \vy^t$ are the input, hidden, memory and output vector respectively. We highlight in blue the memory component and in green the functional component. The edge with the small square represents time delay.} \label{fig:lmn}
\end{figure}
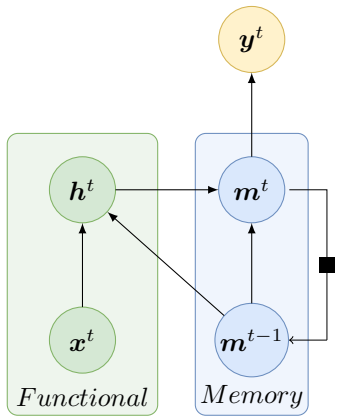
where $N_x,\ N_h,\ N_m$ are the input size, hidden size and memory size respectively, $\mW^{xh} \in \mathbb{R}^{N_h \times N_x}$, \mbox{$\mW^{mh} \in \mathbb{R}^{N_h \times N_m}$}, $\mW^{hm} \in \mathbb{R}^{N_m \times N_h}$, $\mW^{mm} \in \mathbb{R}^{N_m \times N_m}$ are the model parameters matrices, and $\sigma$ is a non-linear activation function ($tanh$ for the purpose of this paper). The output of the memorization component $\vm^t$ is used as the output of the entire network, which can be used to predict the desired output $\vy^t$, for example by using a linear layer:
\begin{equation}\label{eq:lmn_readout}
    \vy^t = \mW^{my}\vm^t.
\end{equation}
Another possibility, previously explored in~\cite{lmn_carta_icann19}, is to use the hidden state $\vh^t$ as the output of the network. We call these two alternatives LMN-m and LMN-h respectively. In this paper we decided to focus on the LMN-m, which we will call LMN from now on, because it showed consistently superior results on all our preliminary experiments.
A schematic view of the LMN is shown in Figure \ref{fig:lmn}. Notice that the linearity of the recurrence does not limit the expressive power of the entire model since given an RNN such that $\vh^t = \sigma(\tilde{\mW}^{xh} \vx^t + \tilde{\mW}^{hh} \vh^{t-1})$, it is possible to initialize an equivalent LMN that computes the same hidden activations by setting $\mW^{xh} = \tilde{\mW}^{xh}$, $\mW^{mh} = \tilde{\mW}^{hh}$, $\mW^{hm} = \mathbb{I}$, $\mW^{mm} = 0$.

The LMN is a direct application of the principles proposed in Section \ref{sec:intro} on the conceptual separation of sequential problems. These principles, applied to the design of recurrent models, help us to define novel architectures and simplify each component without losing expressivity. The linearity of the memorization component allows to control the short-term memorization properties of the entire model in an easy way. For example, we can exploit the linearity of the matrix $\mW^{mm}$ of the LMN to control and adjust the optimal short-term memory capacity\cite{Ganguli2008a}. One advantage of the separation between the two modules is that we can train the memory component separately to encode the sequence of hidden features computed by the functional component. Since our memory module is a LAES, in Section \ref{sec:lmn_pretrain} we will show how to train it using the LAES training algorithm. The proposed approach provides a simple solution to the memorization problem by initializing the memorization component with the solution of the LAES shown in Section \ref{sec:laes}.

\subsection{Model Architecture with MultiScale LAES Memory} \label{sec:ms_lmn}
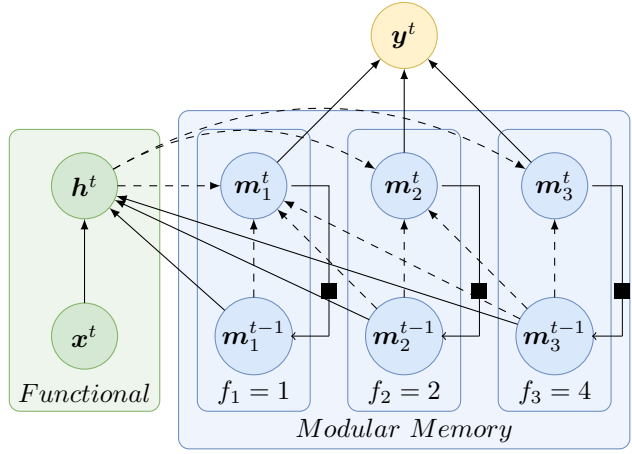
\begin{figure}
    \centering
    \input{img/ms_lmn.tex}
    \caption{Architecture of the MultiScale Linear Memory Network with $3$ memory modules. Dashed connections are affected by the subsampling, while connections with the small square represent time delay.} \label{fig:ms_lmn}
\end{figure}

The memorization of long sequences can be a difficult problem. Since the memory of the LMN is based on an autoencoder with a fixed memory size, it can only encode a finite amount of information, and it could incur in large reconstruction errors when dealing with long sequences. Ideally, a good recurrent model should be able to capture both short-term and long-term dependencies with a small memory size. To address this problem, we extend the LMN with a modular memorization component, divided into separate modules, each one responsible to process the hidden state sequence at a different timescale, as shown in Figure \ref{fig:ms_lmn}. The modules responsible for longer timescales subsample the hidden states sequence to focus on long-term interactions in the data and ignore the short-term ones. In practice, the memorization component is separated into $g$ different modules with exponentially longer sampling rates $1, 2, ..., 2^g$. The connections affected by the subsampling are shown with dashed edges in Figure \ref{fig:ms_lmn}. Notice that, given a maximum dependency length $l_{max}$, the maximum number of different frequencies is $g = \lfloor \log{l_{max}} \rfloor$. Therefore, the memorization component only requires a small number of modules even for long sequences. Each memory module is connected only to slower modules, and not vice-versa, to avoid interference of the faster modules with the slower ones. The organization of the memory into separate modules with a different sampling rate is inspired by the Clockwork RNN\cite{clockwork_koutnik14}, which is an RNN with groups of hidden units that work with different sampling frequencies. Differently from the Clockwork RNN, we apply this decomposition only to the memorization component of the LMN, and not to the hidden units of the functional component. Furthermore, we can exploit the linearity of the memory to achieve better memorization of long sequences, as we will see in the experimental results in Section \ref{sec:experiments}.
In Section \ref{sec:lmn_train}, we will show how to train each memory module to encode the subsampled sequences of hidden activations.

The model update computes at each timestep $t$ an hidden state $\vh^t$ and $g$ memory states $\vm_1^t, \hdots, \vm_g^t$ as follows:
\begin{eqnarray}
 \vh^t &=& \sigma(\mW^{xh} \vx^t + \sum_{i=1}^{g} \mW^{m_i h} \vm^{t-1}_i) \label{eq:ms_lmn_h} \\
 \vm^t_k &=& \begin{cases}
 \vm_{new}^t & t \mod{2^{k-1}} = 0 \\
 \vm_k^{t-1} & otherwise
 \end{cases} \forall k \in 1, \hdots, g \label{eq:ms_lmn_m} \\
 \vm_{new}^t &=& \mW^{h m_k} \vh^t + \sum_{i=1}^{k} \mW^{m_i m_k} \vm^{t-1}_i
\end{eqnarray}
where $\vx^t \in \mathbb{R}^{N_x}$, $\vh^t \in \mathbb{R}^{N_h}$, $\vm_k^t \in \mathbb{R}^{N_m}$. The subsampling of the hidden state sequence is performed by choosing when to update the memory state using the modulo operation. The output can be computed from the output modules as follows:
\begin{equation} \label{eq:ms_lmn_out}
    \vy^t = \sum_{i=1}^g \mW^{m_i y} \vm_i^t.
\end{equation}
Figure \ref{fig:ms_lmn} shows a schematic view of the architecture.
For a more efficient implementation, more amenable to parallel architectures like GPUs, we can combine all the operations performed by Equations \ref{eq:ms_lmn_h}-\ref{eq:ms_lmn_out} for each module into a single matrix multiplication. In the following, we show the procedure for Equation \ref{eq:ms_lmn_m} since the same approach can be applied to Equations \ref{eq:ms_lmn_h} and \ref{eq:ms_lmn_out}\footnote{The equations are shown in the supplementary material.}. First, we notice that the memory modules are ordered by frequency, from fastest to slowest, and their sampling frequencies are powers of $2$. As a consequence, if module $i$ is active, then all the modules $j$ with $j < i$ are also active since $t \mod 2^i = 0$ implies $t \mod 2^j = 0$ whenever $j \geq 0$ and $j < i$. Therefore, we only need to find the maximum index of the active modules $i_{max}^t = max \{ i\ |\ t \mod 2^{i-1} = 0\ \land\ i \leq g \}$ to know which memory modules must be updated. We can combine the activations and parameters of the modules together as follows:
\begin{align*}
    \vm^t = \begin{bmatrix} {\vm_1^t}^\top \\ \vdots \\ {\vm_k^t}^\top \end{bmatrix}, \mW^{hm} = \begin{bmatrix} \mW^{h_1 m} \\ \vdots \\ \mW^{h_g m} \end{bmatrix} \\
    \mW^{mm} = \begin{bmatrix}
        \mW^{m_1 m_1} & & \hdots & \mW^{m_g m_1} \\
        0 & \mW^{m_2 m_2} & \hdots & \mW^{m_g m_2} \\
        \vdots & \ddots & \ddots & \vdots \\
        0 & \hdots & 0 & \mW^{m_g m_g}
    \end{bmatrix}
\end{align*}
Equation \ref{eq:ms_lmn_m} becomes:
\begin{align}
    \vm^t[:i_{max}] &= \mW^{hm}[:i_{max}] \vh^t[:i_{max}] + \mW^{mm}[:i_{max}]\vm^t[:i_{max}] \label{eq:mslmn_m_block1}\\
    \vm^t[i_{max}:] &= \vm^{t-1}[i_{max}:]. \label{eq:mslmn_m_block2}
\end{align}
Please refer to Section \ref{sec:notation} for the usage of the slicing operator $\mW[start:end]$.
Using Eq. (\ref{eq:mslmn_m_block1})-(\ref{eq:mslmn_m_block2}), the subsampling is performed by finding $i_{max}$ which determines which slice of $\vm^t$ must be updated or remain constant. Notice that $\mW_{mm}$ is a block diagonal matrix, therefore it has fewer parameters than a corresponding LMN with $gN_m$ hidden units. Figure \ref{fig:cwrnn-blocks} shows the block structure of the MS-LMN parameters for the memory state update, with the darker blocks being the currently active modules.

\input{img/mslmn_blocks.tex}

\section{Training the Memorization Component} \label{sec:lmn_train}
The separation of sequential processing into two subtasks led us to the architectural separation into two modules, the functional and memorization component of the LMN and MS-LMN described in Section \ref{sec:lmn}. In this section, we study how to exploit the conceptual separation during training.

Given a trained memorization component, the functional subtask requires finding a good hidden representation given the current input element and the memory state. A feedforward network is sufficient to model it making it is relatively easy to solve. The memorization problem, instead, is a recurrent problem, which in general makes it more difficult to be addressed. Therefore, we are interested in finding an approach that allows to easily solve the memorization problem. As a first step toward this goal, the LMN uses a linear recurrence, which simplifies the architecture of the memorization component.

However, if the LMN is trained end-to-end with gradient descent we do not achieve a separation between the two subtasks since the gradient descent will try to solve both subtasks concurrently. In general, we do not expect large benefits by training an LMN by gradient descent instead of a vanilla RNN. Instead, we propose to train the memorization component separately from its functional counterpart to solve the memorization subtask. In this section, we show how to train the memorization component of the LMN to encode the hidden states $\vh^t$ computed by the functional component. The encoding $\vm^t$ can be used to reconstruct the sequence of hidden activations and therefore it represents a compact representation of the entire sequence. The functional component can use this representation to compute the hidden activations based on the entire history by the decoding $\vm^t$.

\subsection{Unrolled Recurrent Model}
\begin{figure}
    \centering
    \input{img/unrolled_rnn.tex}
    \caption{Architecture of the Unrolled RNN. Notice that the memory tape stores only the last $k$ hidden activations.}
    \label{fig:unrolled_rnn}
\end{figure}
The most straightforward implementation of an explicit memory model is a tape delay line, which we will call Unrolled Recurrent Model (URNN) to make explicit its relationship with RNN and LMN, which will be formalized in Section \ref{sec:lmn_equivalence}. The tape memorizes the most recent $k$ hidden states and the URNN uses direct connections between all of them to update the hidden state as follows:
\begin{eqnarray}
    \vh^t & = & \sigma(\mW^{xh}\vx^t + \sum_{i=1}^{k} \mW^{hh}_{i}\vh^{t-i}), \quad t=1, \hdots, k \label{ulm:1}\\
    \vy^t & = & \sigma(\sum_{i=0}^{k} \mW^{hy}_i \vh^{t-i}),\label{ulm:2}
\end{eqnarray}
where $\mW^{hh}_{i} \in \mathbb{R}^{N_h \times N_h}$ represents direct connections between the current hidden state $\vh^t$ and the hidden state at time $t-i$, while $\mW^{hy}_i \in \mathbb{R}^{N_y \times N_h}$ represents the connections between the current output $\vy^t$ and the hidden state at time $t-i$.
The URNN solves the memorization subtask with a lossless finite memory of size $k$. Unfortunately, this model has several disadvantages. First, the number of parameters and its computational cost grows linearly with the size of the tape $k$. Furthermore, its memory is limited to the last $k$ hidden states. Therefore, the memorization of long sequences is easy to achieve with a long delay line but expensive to train due to the parameter explosion. Instead, we would like to encode the entire sequence with a fixed cost in time and space resources. In the next sections we will show how the LMN can solve both of these limitations.

\subsection{Equivalence Between the URNN and the LMN}\label{sec:lmn_equivalence}
\begin{figure}
    \centering
    \input{img/lmn_train.tex}
    \caption{LMN initialization. Notice how the memory state $\vm^t$ is used to reconstruct the memory tape of the URNN. The connection with the small square represents time delay.}\label{fig:lmn_pretrain}
\end{figure}
Using the LMN we can approximate the URNN by replacing the fixed memory of the tape delay line with an autoencoder, which can be used to encode and reconstruct the previous hidden activations. Differently from the tape delay line, the resulting model has an memory equal to the maximum length of the training sequences and can encode sequences of any length with a fixed number of parameters. In the remainder of this section we show how to build an equivalent LMN given a trained URNN.

The first step to build the approximation of the URNN is the collection of the sequence of hidden activations $\vh^1, \hdots, \vh^l$ for each training sample. Given the dataset of the hidden activation's sequences, we can train a LAES to encode the sequences using the algorithm provided in Section \ref{sec:laes}. The obtained model can be used to encode and decode the hidden activations as follows:
\begin{align*}
    \vm^t = \mA\vh^t + \mB\vm^{t-1}  \\
    \begin{bmatrix} \tilde{\vh}^t \\ \tilde{\vm}^{t-1} \end{bmatrix} = \begin{bmatrix} \mA^\top \\ \mB^\top \end{bmatrix}\vm^t.
\end{align*}

The trained LAES can be used to reconstruct the last hidden state of the URNN and the tape delay line by decoding the last $k + 1$ hidden states. Figure \ref{fig:lmn_pretrain} shows the resulting LMN, highlighting the encoding and decoding operations implicit in the connections between the functional and memorization component. Given
\begin{equation*}
    \mU_k = \begin{bmatrix} \mA^\top \\ \mA^\top\mB^\top \\ \vdots \\ \mA^\top{\mB^{k-1}}^\top \end{bmatrix}
\end{equation*}
we can decode the last $k$ hidden activations as follows:
\begin{equation} \label{eq:unrolled_decoder}
    \mU_k \vm^t = \begin{bmatrix} \mA^\top \\ \mA^\top\mB^\top \\ \vdots \\ \mA^\top{\mB^{k-1}}^\top \end{bmatrix} \vm^t = \begin{bmatrix} \tilde{\vh}^t \\ \tilde{\vh}^{t-1} \\ \hdots \\ \tilde{\vh}^{t-k} \end{bmatrix}
\end{equation}
where $\tilde{\vh}^1, \hdots, \tilde{\vh}^{t-k}$ is the reconstruction of the hidden state sequence up to time $t-k$ provided by the LAES. Equation \ref{eq:unrolled_decoder} can be used to approximate a tape delay line. We can control the quality of the approximation by adjusting the number of units of the memory state $\vm^t$ to reduce the reconstruction error. Notice that due to the linearity of the decoder, the decoding of $\vh^{t-j}$ can be done with a single matrix multiplication $\mA^\top {\mB^{j-1}}^\top$.

Using Equation \ref{eq:unrolled_decoder} we can initialize an LMN to approximate a URNN. The LMN $\mW^{xh}$ matrix is initialized using the corresponding matrix $\mW^{xh}$ of the unrolled model. The parameters of the memorization component $\mW^{hm}$ and $\mW^{mm}$ are obtained from the LAES parameters, while the parameters of the functional component can be obtained by combining the decoding operation (Eq. \ref{eq:unrolled_decoder}) with the corresponding parameters of the URNN. The resulting LMN is initialized as follows:
\begin{eqnarray*}
\mW^{hm} & = & \mA \label{eq:pret}\\
\mW^{mm} & = & \mB \\
\mW^{mh} & = & \begin{bmatrix} \mW^{hh}_1 & \hdots & \mW^{hh}_k \end{bmatrix} \mU_k \\
\mW^{mo} & = & \begin{bmatrix} \mW^{hy}_0 & \hdots & \mW^{hy}_k \end{bmatrix} \mU_{k+1}.
\label{eq:pret2}
\end{eqnarray*}
The memory update of the LMN requires a single matrix multiplication, and therefore it is more efficient than the equivalent URNN, which requires $k$ matrix multiplications. The resulting model is also more expressive, since it can memorize longer sequences, up to the maximum length of the sequences in the training set.

Notice that we could the same reconstruction by using an RNN with saturating nonlinearities instead of the LMN, by ignoring the activation function and incurring in some approximation error. However, the nonlinearity inside the recurrence would quickly generate a large approximation error during the encoding process. Instead, the reconstruction shown here for a linear memorization component is exact, except for the reconstruction error of the LAES, which can be minimized by increasing the memory size.

\subsection{Training the LMN for Explicit Memorization}\label{sec:lmn_pretrain}
The equivalence result shows how to construct an equivalent LMN given an unrolled model with a maximum dependency length $k$. We can exploit this result to train the memorization component of the LMN by dividing training into three phases:
\begin{description}
    \item[Unrolled RNN training] The unrolled model is trained to solve the desired sequential problem. The training of this model provides an initialization for the hidden state sequences computed by the functional component.
    \item[LMN initialization] An equivalent LMN is constructed from the trained URNN. This initialization is used to train the memorization component to encode the hidden activations using the LAES training algorithm.
    \item[LMN finetuning] The initialized LMN is finetuned with gradient descent. The memory provided by the LAES encoding can be used to learn a better representation and improve the final performance. This phase trains the functional and memorization component concurrently.
\end{description}

Notice that after the finetuning phase the memory component will not be an autoencoder anymore, i.e., in general, it will not be possible to reconstruct the sequence of hidden activations given $\vm^t$. Nonetheless, the initial encoding learned during the initialization provides a good initial representation to solve tasks that require the memorization of long sequences.
Algorithm \ref{alg:lmn_train} shows the pseudocode for the entire procedure.
\input{pcode/pseudo_train_lmn.tex}

\subsection{Incremental Training for MultiScale-LMN}\label{sec:mslmn_pretrain}

\begin{figure}
    \centering
    \input{img/ms_lmn_init.tex}
    \caption{MS-LMN pretraining.}
    \label{fig:mslmn_pretrain}
\end{figure}
The LMN initialization adopted in the previous section is effective whenever the memory state $\vm^t$ is large enough to encode a good approximation of the entire sequence. However, problems that contain extremely long sequences may require a large memory. The modular memory of the MS-LMN is more effective in this situation since we can use each module to provide an approximation of the subsampled sequence. Therefore, we would like to train the memory modules of the MS-LMN to solve the memorization problem by encoding the subsampled hidden activation sequences.
The training algorithm for the MS-LMN works incrementally: at first, a single LMN is trained, possibly trained with the algorithm shown in Section \ref{sec:lmn_train}. After a fixed number of epochs, an additional memory module with a slower frequency is added and it is initialized to encode the hidden activations of the current model. The resulting model is then trained with SGD and the process is repeated until all modules have been added.

The addition of a new module works as follows. First, let us assume to have already trained an MS-LMN with $g$ modules as defined in Equation \ref{eq:ms_lmn_h},\ref{eq:ms_lmn_m}. We collect the sequences of hidden activations $\vh^1, \hdots, \vh^t$ for each sample in the training set, subsample them with frequency $2^g$, and train a LAES with parameters $\mA_{g+1}$ and $\mB_{g+1}$. A new module with sampling frequency $2^{g}$ is added to the previous model and the new connections are initialized as follows:
\begin{align*}
    \mW^{h m_{g+1}} = \mA_{g+1} \\
    \mW^{m_{g+1} m_{g+1}} = \mB_{g+1} \\
    \mW^{m_{g+1} o} = \mW_{g+1},
\end{align*}
where $\mW_{g+1}$ is obtained by training a linear model to predict the desired output $\vy^t$ from the memory state of the entire memory $\vm_1^t, \hdots, \vm_{g+1}^t$. The remaining new connections can be initialized with zeros. After the initialization of the new connections, the entire model is trained end-to-end by gradient descent. Notice that during this phase the entire model is trained, which means that the connections of the old modules can still be updated. This process is repeated until all the memory modules have been added and the entire model has been trained for the last time. Algorithm \ref{alg:ms_lmn_train} shows the pseudocode for the entire training procedure.

The proposed approach provides several advantages by combining the strength of the modular memorization component of the MS-LMN with the autoencoding of the LMN while tackling some of the shortcomings of the LMN training algorithm.
To train the LMN memorization component, it is necessary to provide a learned representation for the functional component activations $\vh^t$, which will be encoded by the LAES. The LMN training algorithm solves this problem by training a separate model, the unrolled model, to initialize the functional component representation. Unfortunately, training the unrolled model entails several drawbacks. The URNN fixes a maximum memory length $k$ and it is expensive to train. As shown in Section \ref{sec:exp_midi}, the unrolled model obtains worse results than an equivalent LMN trained from scratch. This is probably due to the difficulty of training such a large overparameterized model using SGD.
The MS-LMN incremental training circumvents this problem by avoiding the unrolled model altogether, by training incrementally the functional component and the newly added modules of the memorization component. The initialization of the new modules directly encodes the activations of the previous MS-LMN instead of using the unrolled model. This allows the network to learn a representation for the hidden features $\vh^t$, which can be improved with the introduction of additional memory modules with a slower frequency, better suited to capture longer dependencies.

\input{pcode/pseudo_train_ms_lmn.tex}

\section{Experiments} \label{sec:experiments}
In this section, we show the results of the experimental assessment of LMN and MS-LMN. We evaluate the models on three different tasks, comprising multiple datasets, that stress the memorization capacity of recurrent models:
\begin{description}
    \item[MIDI Music Modeling] Polyphonic music modeling with MIDI datasets using a piano-roll representation.
    \item[Sequence Generation] A synthetic dataset that requires the model to generate a long audio signal without any external input.
    \item[Common Suffix TIMIT] Word classification using speech signals, restricted to classes of words that share a long common suffix.
\end{description}

Music modeling is a good task to test a memorization model due to the repeated patterns typical of music scores, which can be repeated in distant timesteps and therefore must be accurately remembered to accurately model the music. The MS-LMN has been designed to capture long-term dependencies in time series signals, whenever it is possible to subsample the data. Therefore, audio signals are a natural choice to test this architecture. For TIMIT, we decided to adopt the experimental setup of \cite{clockwork_koutnik14}, which selects from the entire dataset words with a common suffix, which means the model must remember the initial part of the sequence to correctly classify the word. We compare against standard recurrent models, such as the Vanilla RNN\cite{elman_rnn} and LSTM\cite{hochreiter_lstm97}, and state-of-the-art models in the benchmarks that we use to evaluate the LMN, such as RNN-RBM\cite{boulanger2012_icml_rnnrbm} for MIDI music modeling, and Clockwork RNN\cite{clockwork_koutnik14} for TIMIT.

The models and experiments are implemented using Pytorch\cite{pytorch_paszke19}. The source code is available online\footnote{\url{https://gitlab.itc.unipi.it/AntonioCarta/mslmn}}.

\subsection{MIDI Music Modeling}\label{sec:exp_midi}
We evaluated the LMN on the problem of music modeling using four datasets of polyphonic music preprocessed in a piano roll representation. Each dataset comprises different styles and different degrees of polyphony~\cite{boulanger2012_icml_rnnrbm}. During the preprocessing, each sequence is sampled at equal timesteps to obtain a feature vector composed of $88$ binary values representing the piano notes from A0 to C8. Each note is set to $1$ if it is currently being played or $0$ if it is not. The task is to predict the notes played at the next timestep given the sequence of previous notes. The performance of each model is evaluated using the frame-level accuracy as defined in \cite{bay_flacc}. We used the same train-validation-test split as in \cite{boulanger2012_icml_rnnrbm}. The datasets have varying degrees of difficulty, ranging from datasets with a homogeneous style, like the Bach chorales or folk music, to heterogeneous repositories of scores composed for piano. The heterogeneity results in widely different accuracies for each dataset. Table \ref{tbl:midi-stats} shows the number of samples and the maximum length of the sequences for each dataset.

\begin{table}[t]
    \centering
    \caption{Number of samples and maximum length for the MIDI music modeling datasets.}\label{tbl:midi-stats}
    \begin{tabular}{lcc}
    \toprule
                  & Samples & max. timesteps \\
    \toprule
    JSB Chorales  & 382       & 160 \\
    MuseData      & 783       & 4273 \\
    Nottingham    & 1037      & 1793 \\
    Piano MIDI    & 124       & 3857 \\
    \bottomrule
    \end{tabular}
\end{table}

For this experiment, we compare two slightly different configurations of the LMN architecture, computing the output using the hidden state $\vh^t$ of the functional component (LMN-h), or using the state $\vm^t$ of the memorization component (LMN-m), as described in Eq. \ref{eq:lmn_readout}. The models have been tested using a random initialization of the parameters. Additionally, we have tested the LMN-m configuration with parameters initialized using separate training for the memorization component, as described in Section \ref{sec:lmn_pretrain}. Since our preliminary experiments showed that the LMN-m configuration always outperforms the LMN-h, in the following experiments we only use this configuration (tagged only as LMN). The LMN results are compared against a number of reference models from the literature. Specifically, we consider an RNN with random initialization or using the pretraining scheme described in \cite{preatrain_la_Pasa2014}, an LSTM network, and the RNN-RBM model (for which we report the original results from \cite{boulanger2012_icml_rnnrbm}). Note that the Nottingham dataset has been expanded since the publication of \cite{boulanger2012_icml_rnnrbm} and therefore the results are not fully comparable.
\begin{table}
    \centering
    \caption{Hyperparameters for the models trained on MIDI Music Modelling tasks.}
    \label{tbl:hp_midi}
    \begin{tabular}{cc}
        \toprule
         Hyperparameter & Grid Values \\
         \midrule
         hidden units $N_h$ &  $50,\ 100,\ 250,\ 500,\ 750$\\
         \midrule
         hidden and memory units $(N_h, N_m)$& \shortstack{$(50, 50), (50, 100),\ (100, 100),\ $ \\ $(100, 250),\ (250, 250),\ (250, 500)$} \\
         \midrule
         l2-decay & $10^{-4},\ 10^{-5},\ 10^{-6},\ 10^{-7},\ 0$ \\
         \bottomrule
    \end{tabular}
\end{table}
\begin{table*}
    \centering
    \caption{Frame-level accuracy computed on the test set for each model. RNN-RBM results are taken from \cite{boulanger2012_icml_rnnrbm}}
    \label{tbl:midi-results}
    \begin{tabular}{lcccc}
    \toprule
                & JSB Chorales & MuseData & Nottingham & Piano MIDI \\
    \toprule
    RNN            & 31.00        & 35.02    & 72.29      & 26.52        \\
    pret-RNN    & 30.55        & 35.47    & 71.70      & 27.31        \\
    LSTM        & 32.64        & 34.40    & 72.45      & 25.08      \\
    RNN-RBM*     & 33.12        & 34.02    & \textbf{75.40}      & \textbf{28.92}      \\\midrule
    LMN-h       & 30.61        & 33.15    & 71.16      & 26.69      \\
    LMN-m       & 33.98        & 35.56    & 72.71      & 28.00      \\
    pret-LMN-m  & \textbf{34.49}        & \textbf{35.66}    & 74.16      & 28.79 \\          
    \bottomrule
    \end{tabular}
\end{table*}
All the networks have been optimized using Adam~\cite{adam_kingma2014} with a fixed learning rate of $0.001$ using early stopping on the validation set to limit the number of epochs.
Except for the RNN-RBM, all the architectures have a single layer. The RNN-RBM has been explicitly designed to solve music modeling problems, while we decided to focus on a generic architecture for sequential problems. For the RNN and LSTM models, we have selected the number of hidden recurrent units, while for the LMN we also evaluated configurations with a different number of functional and memory units. All models have been regularized using L2 weight decay. A detailed list of the hyperparameters explored by the grid search is shown in Table \ref{tbl:hp_midi}.

The URNN used to initialize the LMN is trained with an unfolding length set to $k=10$, with hidden sizes equal to the corresponding LMN. The unfolding length was chosen to guarantee a good tradeoff between the memory required to run the model and the expressiveness of the model since in principle we would always prefer a large unfolding length. In practice, we also noticed that for large $k$ the model becomes more difficult to train and achieves a worse performance. The models use a $tanh$ activation function for the hidden units and a sigmoid activation for the outputs.

The test performances for the best configuration of each model (selected on the validation set) are reported in Table \ref{tbl:midi-results}. The results show that the LMN-m is competitive with gated recurrent architectures like the LSTM, even without training the memorization component separately. The LMN obtains better results in two different datasets when compared to the RNN-RBM, despite being a shallower model. On the Piano MIDI dataset, the RNN-RBM is slightly better than the LMN, while the Nottingham dataset is tested using the updated version, and therefore the results are not exactly comparable. The RNN has also been evaluated using a pretraining scheme (note that RNN-RBM uses pretraining as well \cite{boulanger2012_icml_rnnrbm}). The training of the memorization component for the LMN is always effective and improves the results on every dataset, while in comparison the pretraining for the RNN seems less effective. As argued at the end of Section \ref{sec:lmn_equivalence}, this is not surprising since the LMN initialization is exact, while the approach proposed in~\cite{preatrain_la_Pasa2014} is only an approximation of the LAES. Furthermore,~\cite{preatrain_la_Pasa2014} only encodes the input sequence, while the LMN encodes the hidden states. LSTM models are not pretrained since the proposed pretraining approach cannot be easily adapted to gating units. We argue that this is an example where it can be appreciated the advantage of dealing with a less complex architecture which, despite its apparent simplicity, leads to excellent performance results.

\subsection{Analysis of the Training for the Memorization Component}
    \begin{figure*}
    \centering
    \resizebox{\textwidth}{!} {
      \begin{subfigure}{.3\textwidth}
        \centering
        \includegraphics[width=.9\linewidth]{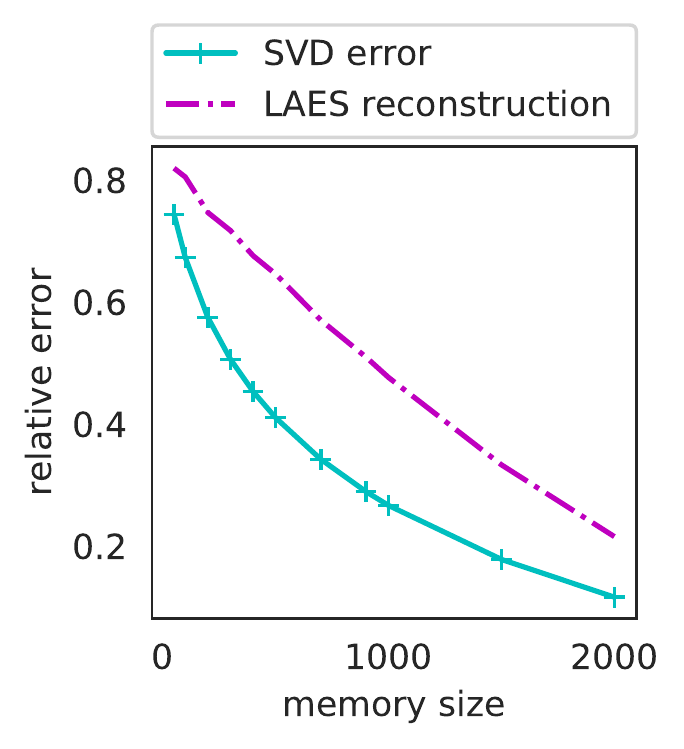}
        \caption{}
        \label{fig:pret-sub1}
      \end{subfigure}%
      \begin{subfigure}{.3\textwidth}
        \centering
        \includegraphics[width=.9\linewidth]{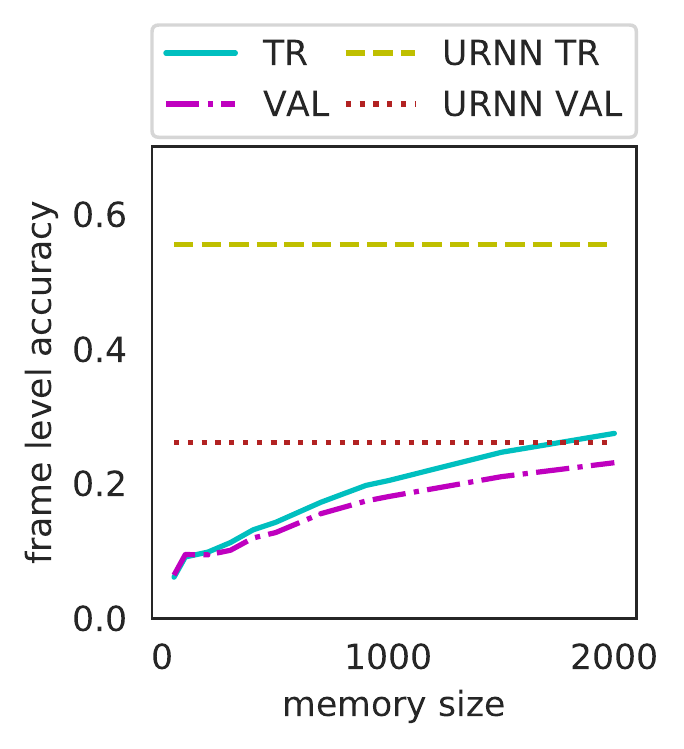}
        \caption{}
        \label{fig:pret-sub2}
      \end{subfigure}
      \begin{subfigure}{.3\textwidth}
        \centering
        \includegraphics[width=.9\linewidth]{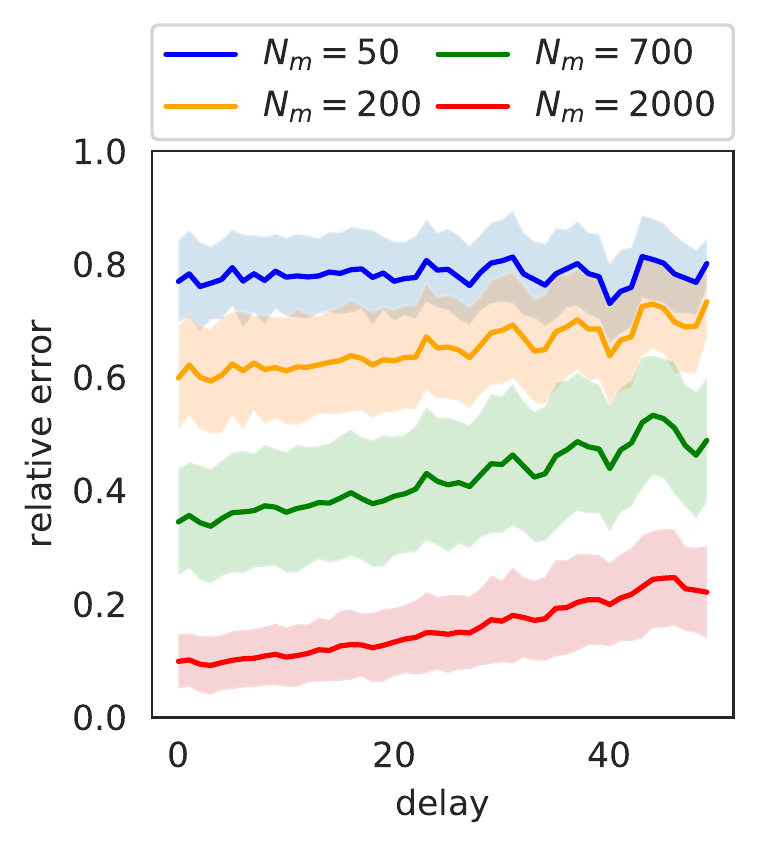}
        \caption{}
        \label{fig:pret-sub3}
      \end{subfigure}
    }
    \caption{Performance of the different training stages computed on the JSB Chorales dataset. Figure \ref{fig:pret-sub1} shows the SVD reconstruction error and the LAES reconstruction error for the training set hidden state sequences. Figure \ref{fig:pret-sub2} shows the pretrained LMN error after the initialization, compared against the URNN on the train (TR) and validation (VAL) sets. Figure \ref{fig:pret-sub3} shows the behavior of the average LAES reconstruction error through time.}
    \label{fig:pretrain}
\end{figure*}
The success of the training algorithm for the memorization component depends on several different properties: the ability of the unrolled model to learn the task, the reconstruction of the sequences of hidden activations performed by the LAES, and the error caused by the approximation during the initialization. In this section we show the effect of each of these components for the LMN trained on the JSB Chorales dataset. Figure \ref{fig:pret-sub1} shows the average reconstruction error made by the SVD factorization of $\mXi$ during the training of the LAES. As shown in Section \ref{sec:laes}, $\mXi$ is the matrix containing the hidden states' subsequences of the unrolled model, computed on the training set. On the same plot, we overlay the reconstruction error of the corresponding trained linear autoencoder (LA). As expected, the reconstruction error steadily decreases for both models as the number of memory units grows. However, we need a large number of memory units to achieve a zero error.

The parameter matrices of the linear autoencoder are used to initialize the LMN. Figure \ref{fig:pret-sub2} shows the performance obtained by the URNN and the corresponding equivalent LMN on the training and validation sets of JSB Chorales. The performance is computed after the initialization with the LAES and before the fine-tuning phase. It can be seen how the LMN performance on the validation set after the initialization is close to that obtained by the URNN, while the LMN greatly reduces the number of parameters used with respect to the URNN configuration. It must also be noted that the performance of the unfolded network is lower than that of a trained LMN model initialization with random values. This result highlights the need for a fine-tuning phase after the pretraining initialization. Furthermore, it justifies further research on alternative methods that do not require the training of the unrolled model, like the incremental training presented in Section \ref{sec:mslmn_pretrain}.

Figure \ref{fig:pret-sub3} shows the average LAES reconstruction error through time. The hidden state sequences are computed using a URNN with $50$ hidden units, while the LAES has a varying number of memory units. The sequences of hidden states are encoded by the LAES for the first $50$ timesteps. Afterward, the last state of the LAES is used to reconstruct the entire sequence. As expected, a larger memory size lowers the reconstruction error. Notice that even the distant hidden states can be reconstructed with a good reconstruction error.

\subsection{Experiments with Modular Memories}

In the following experiments, we study the ability of the modular memorization component of the MS-LMN to capture long-term dependencies in two different sequential problems. The model is compared against the LMN and the CW-RNN. Each model is a single layer recurrent neural network with \emph{tanh} activation function (only in the functional component for LMN and MS-LMN). The number of parameters is chosen by model selection from a fixed list of options. Table \ref{tab:num-params} shows the number of parameters and hidden units for several configurations. Notice that given a fixed number of parameters, for LMN, CW-RNN, and MS-LMN, it is possible to choose different architectures by varying the number of hidden units, memory units, and memory modules. We tested the MS-LMN on audio signals since they are a natural choice to assess the modular memory, which works by subsampling its input sequence. The first task, \emph{Sequence Generation}, is a synthetic problem that requires the model to output a given signal without any eternal input. The second task, \emph{Spoken Word Classification}, is a sequence classification task that uses a subset of spoken words extracted from TIMIT\cite{garofolo_timit_data93}. The task is designed to have long-term dependencies by only considering a restricted subset of words that have a common suffix. Throughout all the experiments we use the Adam optimizer~\cite{adam_kingma2014} with L2 weight decay.

\subsection{Sequence Generation}

\begin{table}[t]
    \centering
    \small
    \caption{Results for the Sequence Generation task. Performance computed using NMSE (lower is better).}\label{tab:gen-hyperparams}
    \begin{tabular}{lccccc}
        \toprule
                            & RNN                    & LSTM          & CW-RNN        & LMN                    & MS-LMN \\
        \midrule
        Hidden Units        & 31                     & 15            & 36            & 2                      & 1 \\
        Memory Units        & /                      & /             & /             & 29                     & 36 \\
        Learning Rate       & \num{1e-03}            & \num{1e-02}   & \num{5e-05}   & \num{5e-04}            & \num{5e-03} \\
        \# parameters       & 1000                   & 1000          & 1000          & 1000                   & 1000 \\
        Epochs              & 6000                   & 12000         & 2000          & 5000                   & 8000 \\
        \midrule
        NMSE ($10^{-3}$)                & 79.5                & 20.7       & 12.5       & 38.4                & \textbf{0.116} \\
        \bottomrule
    \end{tabular}
\end{table}

\begin{table}[tb]
    \centering
    \small
    \caption{Number of parameters and corresponding number of hidden units for different configurations trained on the Sequence Generation task. For CW-RNN and MS-LMN we show the total number of units and the number for each module (the latter between parenthesis).}\label{tab:num-params}
    \begin{tabular}{cccccc}
        \toprule
        \#parameters & RNN   & LSTM  & CW-RNN     & LMN        & MS-LMN \\
                     & $N_h$ & $N_h$ & $N_h, (\frac{N_h}{g})$ & $N_h, N_m$ & $N_h, N_m, (\frac{N_m}{g})$ \\
        \midrule
        100  & 9  & 4  & 9, (1)  & 4, 6   & 1, 9, (1) \\
        250  & 15 & 7  & 18, (2) & 7, 10  & 1, 18, (2) \\
        500  & 22 & 10 & 27, (3) & 11, 13 & 1, 27, (3) \\
        1000 & 31 & 15 & 36, (4) & 2, 29 & 1, 36, (4) \\
        10000         & 82        & 40         & 84, (12)     & 20, 71    & 15, 77, (11) \\
        \bottomrule
    \end{tabular}
\end{table}
The Sequence Generation task is a synthetic problem that requires the model to output a given signal without any eternal input. We extracted a sequence of 300 data points generated from a portion of a music file, sampled at 44.1 KHz starting from a random position. Sequence elements are scaled in the range \( [-1,1] \).
The task stresses the ability to learn long-term dependencies by requiring the model to encode the entire sequence without any external input.
The CW-RNN\cite{clockwork_koutnik14} is the state-of-the-art on this task, where it reaches better results than comparable RNNs and LSTMs.

We tested each model with $4$ different numbers of parameters in $\{100, 250, 500, 1000\}$, by varying the number of hidden neurons. The models are trained to optimize the Normalized MSE. Table~\ref{tab:gen-hyperparams} shows the most relevant hyperparameters of the best configuration for each model which were found with a random search. The CW-RNN and MS-LMN use $9$ modules with exponential clock rates \( \{1,2,4,8,16,32,64,128,256\} \). Notice that adding more modules would be useless since the sequence length is $300$. The number of hidden units for the CW-RNN and of memory units for the MS-LMN is the number of modules times the number of units per module. For the LSTM, we obtained the best results by initializing the forget gate to $5$, as suggested in \cite{Cummins2000}.

The results of the experiments are reported in Table~\ref{tab:gen-hyperparams}. Figure~\ref{fig:gen-plots} shows the reconstructed sequence for each model. The results confirm those found in \cite{clockwork_koutnik14} for the CW-RNN and show a net advantage of the CW-RNN over the RNN and LSTM. The LMN provides an approximation of the sequence, slightly worse than that of the LSTM, which closely follows the global trend of the sequence, but it is not able to model small local variations. The MS-LMN obtains the best results and closely approximates the original sequence. The model combines the advantages of a linear memory, as shown by the LMN performance, with the hierarchical structure of the CW-RNN. The result is that the MS-LMN is able to learn both the short-term and long-term variations of the signal.

\begin{figure}
    \centering
    \input{img/gen_seq.tex}
    \caption{Generated output for the Sequence Generation task. The original sequence is shown with a dashed blue line while the generated sequence is a solid green line.}\label{fig:gen-plots}
\end{figure}

\subsection{Common Suffix TIMIT}
\begin{table}[t]
    \centering
    \caption{Test set accuracy on Common Suffix TIMIT. Variance computed by training with different random seeds for $5$ times using the best hyperparameters found during the model selection.}\label{tbl:timit_results}
    \begin{tabular}{llll}
    \toprule
    Model                    &$n_h$ &$n_m$ & Accuracy    \\
    \midrule
    LMN (h=52)               & 52   & -    & 55.0 +-1.0  \\
    CW-RNN (h=13,modules=7)  & 13   & -    & 74.4 +- 2.9 \\
    MS-LMN (h=25, m=25)      & 25   & 25   & 78.0+-3.4   \\
    pret-MS-LMN (h=25, m=25) & 25   & 25   & 79.6+-3.8   \\
    \bottomrule
    \end{tabular}
\end{table}

TIMIT~\cite{garofolo_timit_data93} is a speech corpus for training acoustic-phonetic models and automatic speech recognition systems. The dataset contains recordings from different speakers in major American dialects, provides a train-test split and information about words and phonemes for each recorded audio file.

Since we are interested in the ability of our model to capture long-term dependencies, we extract from TIMIT a subset of words that have a common suffix, following the preprocessing of \cite{clockwork_koutnik14} as follows. We took $25$ words pronounced by $7$ different speakers, for a total of $175$ examples. The chosen words can be categorized into $5$ clusters based on their suffix:
\begin{itemize}
    \item \emph{Class 1}: making, walking, cooking, looking, working
    \item \emph{Class 2}: biblical, cyclical, technical, classical, critical
    \item \emph{Class 3}: tradition, addition, audition, recognition, competition
    \item \emph{Class 4}: musicians, discussions, regulations, accusations, conditions
    \item \emph{Class 5}: subway, leeway, freeway, highway, hallway
\end{itemize}

The common suffix makes the task more difficult since the model is forced to remember the initial part of the sequence to correctly classify each word.

Each file contained in the TIMIT dataset is a WAV containing the recording of a sentence, therefore we trimmed it to select a single word using the segmentation metadata provided with the dataset. The words extracted from this procedure are preprocessed to extract a sequence of MFCC coefficients using a window length of 25ms, a step of 1ms, preemphasis filter of 0.97, 13 cepstrum coefficients, where we replaced the zeroth cepstral coefficient with the \emph{energy} (the log of total frame energy). As a result, we obtain $13$ features for each timestep. We normalized each feature to have mean $0$ and variance $1$.

To allow a direct comparison with the work in~\cite{clockwork_koutnik14}, we split the dataset by taking 5 words for training and 2 for test from each class. This split ensures a balanced train and test set. During training we added gaussian noise to the sequence with a standard deviation of $0.6$. Due to the small size of the dataset, we do not use an additional train-validation split and we use the clean version of the training set as a validation set, as done in \cite{clockwork_koutnik14}.
Unfortunately, \cite{clockwork_koutnik14} does not provide the exact split used in their experiments. Given the small size of the dataset, in our experiments we have found a great variance between different splits. Therefore, we cannot directly compare against the results in \cite{clockwork_koutnik14} and we decided to train the CW-RNN with our train-test split. To ensure the reproducibility of our work and a fair comparison for future work, we provide the splits in the supplementary material.

We found the best hyperparameters with a random search on the batch size in $\{1, 25\}$, l2 decay in $\{0, 10^{-3}, 10^{-4}, 10^{-5}\}$, learning rate in $\{10^{-3}, 10^{-4}\}$ and hidden units per module in $[5, 40]$. Each model is trained to minimize the cross-entropy loss. When using a batch size equal to \( 25 \) we keep the classes balanced by taking one sample per class. The CW-RNN and MS-LMN use 7 modules with exponential clock rates \( \{1,2,4,8,16,32,64\} \), since the longer sequence has 97 data points. We initialize the LSTM forget gate to $5$, as suggested in \cite{Cummins2000}.

Table \ref{tbl:timit_results} shows the results of the experiment. The CW-RNN and MS-LMN obtain much better results than the LMN due to their ability to learn long-term dependencies. The MS-LMN shows superior results compared to the CW-RNN, and the incremental pretraining improves the performance of the LMN substantially.

\section{Conclusion} \label{sec:conclusion}

In this paper, we proposed a conceptual separation of sequential processing into two different subtasks, the functional and memorization subtasks. By focusing on the memorization problem, which is the recurrent part of the problem, we proposed two novel recurrent neural networks and corresponding training algorithms. We exploited the architectural separation to design two specialized algorithms to train the memorization component, one for each architecture. The algorithms train the memorization component to encode the hidden activations computed by the functional component. The memory state can be used by the functional component to reconstruct its previous activations. The experimental results show that solving the memorization problem with these specialized algorithms improves the performance on several benchmarks. In general, the approach seems useful to solve problems that require to memorize long sequences and remember them after a large number of timesteps.

In the future, we plan to extend these approaches to alternative memorization models and training algorithms for the memorization component. For example, the subtasks separation can be applied to tree or graph-structured data to design novel models.
Another open question is how to generalize this approach to more complex memorization components, like the associative memories used in MANN models, and different sequential tasks, like natural language translation, question answering, and other complex sequential tasks.

Finally, it must be noted that the current approach does not completely separate the training of the two components. In particular, during the finetuning phase of the LMN and the end-to-end training for the MS-LMN, the two components are trained concurrently. In the future, it would be interesting to explore the possibility to completely separate the training of the two components during the entire training process.

%% file: img/fun_mem_arch.tex
\definecolor{blue_light}{RGB}{218,232,252}
\definecolor{blue_dark}{RGB}{108,142,191}
\definecolor{green_light}{RGB}{213,232,212}
\definecolor{green_dark}{RGB}{130,179,102}
\definecolor{yellow_light}{RGB}{255,242,204}
\definecolor{yellow_dark}{RGB}{214,182,86}

\begin{tikzpicture}
    \tikzstyle{sty_memory} = [draw=blue_dark,fill=blue_light,outer sep=0,inner sep=1,minimum size=50]
    \tikzstyle{sty_functional} = [draw=green_dark,fill=green_light,outer sep=0,inner sep=1,minimum size=50]
    \tikzstyle{sty_out} = [draw=yellow_dark,fill=yellow_light,circle,outer sep=0,inner sep=1,minimum size=40]
    \tikzstyle{myedgestyle} = [-latex]
    \tikzstyle{sty_module} = [rounded corners,fill opacity=0.4]

    
    \node[sty_out] (y) at (-1, 1) {$Output$};
    \node[sty_functional] (f) at (-1, -1) {$Functional$};
    \node[sty_memory] (m) at (1.5, -1) {$Memory$};
    \node[sty_out] (x) at (-1, -3) {$Input$};
    
    \draw [myedgestyle] (x) edge (f);
    \draw [myedgestyle] (f) edge (m);
    \draw [myedgestyle] (m) edge (f);
    \draw [myedgestyle] (f) edge (y);
    \draw [myedgestyle] (m) edge (y);
    \draw[myedgestyle] (m) to [out=0,in=-90,loop,looseness=4.8] (m);
\end{tikzpicture}

%% file: img/lmn.tex
\definecolor{blue_light}{RGB}{218,232,252}
\definecolor{blue_dark}{RGB}{108,142,191}
\definecolor{green_light}{RGB}{213,232,212}
\definecolor{green_dark}{RGB}{130,179,102}
\definecolor{yellow_light}{RGB}{255,242,204}
\definecolor{yellow_dark}{RGB}{214,182,86}

\begin{tikzpicture} 
 
    \tikzstyle{sty_memory} = [draw=blue_dark,fill=blue_light,circle,outer sep=0,inner sep=1,minimum size=25]
    \tikzstyle{sty_functional} = [draw=green_dark,fill=green_light,circle,outer sep=0,inner sep=1,minimum size=25]
    \tikzstyle{sty_out} = [draw=yellow_dark,fill=yellow_light,circle,outer sep=0,inner sep=1,minimum size=25]
    \tikzstyle{myedgestyle} = [-latex]
    \tikzstyle{sty_module} = [rounded corners,fill opacity=0.4]

    \draw [draw=green_dark,fill=green_light,sty_module] (-2.25,-2) rectangle (-0.25,1.75);   
    \node[sty_functional] (x) at (-1.25, -1) {$\vx^t$};
    \node[sty_functional] (h) at (-1.25, 1) {$\vh^t$};
    \draw [myedgestyle] (x) edge (h);
    \node[sty_out] (y) at (1, 3) {$\vy^t$};

    \draw [draw=blue_dark,fill=blue_light,sty_module] (0.25,-2) rectangle (1.75,1.75);
    \node[sty_memory] (m) at (1, 1) {$\vm^t$};
    \node[sty_memory] (mp) at (1, -1) {$\vm^{t-1}$};
    \draw [myedgestyle] (h) edge (m);
    \draw [myedgestyle] (mp) edge (m);
    \draw [myedgestyle] (mp) edge (h);
    \draw [myedgestyle] (m) edge (y);
    
    \draw[very thin] (1.5, 1) -- (2, 1) -- (2, -1);
    \draw[very thin, ->] (2, -1) -- (1.5, -1);
    \fill [black] (1.9,-0.1) rectangle (2.1,0.1);

    \node[] (f_text) at (-1.25, -1.75) {$Functional$};
    \node[] (f_text) at (1, -1.75) {$Memory$};
\end{tikzpicture}

%% file: img/ms_lmn.tex
\definecolor{blue_light}{RGB}{218,232,252}
\definecolor{blue_dark}{RGB}{108,142,191}
\definecolor{green_light}{RGB}{213,232,212}
\definecolor{green_dark}{RGB}{130,179,102}
\definecolor{yellow_light}{RGB}{255,242,204}
\definecolor{yellow_dark}{RGB}{214,182,86}

\begin{tikzpicture}

    \tikzstyle{sty_memory} = [draw=blue_dark,fill=blue_light,circle,outer sep=0,inner sep=1,minimum size=25]
    \tikzstyle{sty_functional} = [draw=green_dark,fill=green_light,circle,outer sep=0,inner sep=1,minimum size=25]
    \tikzstyle{sty_out} = [draw=yellow_dark,fill=yellow_light,circle,outer sep=0,inner sep=1,minimum size=25]
    \tikzstyle{myedgestyle} = [-latex]
    \tikzstyle{sty_module} = [rounded corners,fill opacity=0.1]
    \tikzstyle{sty_module_all} = [rounded corners,fill opacity=0.4]

    \draw [draw=green_dark,fill=green_light,sty_module_all] (-2.25,-2) rectangle (-0.25,1.75);   
    \node[sty_functional] (x) at (-1.25, -1) {$\vx^t$};
    \node[sty_functional] (h) at (-1.25, 1) {$\vh^t$};
    \draw [myedgestyle] (x) edge (h);
    \node[sty_out] (y) at (3, 3) {$\vy^t$};

    \draw [draw=blue_dark,fill=blue_light,sty_module_all] (0.,-2.5) rectangle (6,2);
    
    \draw [draw=blue_dark,fill=blue_light,sty_module] (0.25,-2) rectangle (1.75,1.75);
    \node[sty_memory] (m) at (1, 1) {$\vm_1^t$};
    \node[sty_memory] (mp) at (1, -1) {$\vm_1^{t-1}$};
    \draw [myedgestyle, dashed] (h) edge (m);
    \draw [myedgestyle, dashed] (mp) edge (m);
    \draw [myedgestyle] (mp) edge (h);
    \draw [myedgestyle] (m) edge (y);
    \node at (1.0, -1.75) {$f_1=1$};

    \draw [draw=blue_dark,fill=blue_light,sty_module] (2.25,-2) rectangle (3.75,1.75);
    \node[sty_memory] (m2) at (3, 1) {$\vm_2^t$};
    \node[sty_memory] (mp2) at (3, -1) {$\vm_2^{t-1}$};
    \draw [myedgestyle, bend left, dashed] (h) edge (m2);
    \draw [myedgestyle, dashed] (mp2) edge (m2);
    \draw [myedgestyle] (mp2) edge (h);
    \draw [myedgestyle] (m2) edge (y);
    \draw [myedgestyle, dashed] (mp2) edge (m);
    \node at (3.0, -1.75) {$f_2=2$};

    \draw [draw=blue_dark,fill=blue_light,sty_module] (4.25,-2) rectangle (5.75,1.75);
    \node[sty_memory] (m3) at (5, 1) {$\vm_3^t$};
    \node[sty_memory] (mp3) at (5, -1) {$\vm_3^{t-1}$};
    \draw [myedgestyle, bend left, dashed] (h) edge (m3);
    \draw [myedgestyle, dashed] (mp3) edge (m3);
    \draw [myedgestyle] (mp3) edge (h);
    \draw [myedgestyle] (m3) edge (y);
    \draw [myedgestyle, dashed] (mp3) edge (m);
    \draw [myedgestyle, dashed] (mp3) edge (m2);
    \node at (5.0, -1.75) {$f_3=4$};
    
    \draw[very thin] (1.5, 1) -- (2, 1) -- (2, -1);
    \draw[very thin, ->] (2, -1) -- (1.5, -1);
    \fill [black] (1.9,-0.5) rectangle (2.1,-0.3);
    
    \draw[very thin] (3.5, 1) -- (4, 1) -- (4, -1);
    \draw[very thin, ->] (4, -1) -- (3.5, -1);
    \fill [black] (3.9,-0.5) rectangle (4.1,-0.3);
    
    \draw[very thin] (5.5, 1) -- (5.9, 1) -- (5.9, -1);
    \draw[very thin, ->] (5.9, -1) -- (5.5, -1);
    \fill [black] (5.8,-0.5) rectangle (6.0,-0.3);

    \node[] (f_text) at (-1.25, -1.75) {$Functional$};
    \node[] (f_text) at (3, -2.25) {$Modular\ Memory$};
\end{tikzpicture}

%% file: img/mslmn_blocks.tex
\begin{figure}[tb]
    \centering
    \resizebox{0.95\linewidth}{!}{
    \begin{tikzpicture}[
        textnode/.style={},
        squareactive/.style={rectangle, draw=black, fill=black!25, minimum size=5mm},
        squareinactive/.style={rectangle, draw=black, fill=black!5, minimum size=5mm},
        squareempty/.style={rectangle, draw=black!0, fill=black!0, minimum size=5mm},
        square/.style={rectangle, draw=black, fill=black!0, minimum size=5mm}
    ]
    
    \begin{scope}[local bounding box=ht]
    \foreach \i in {0,...,4} {
        \ifthenelse{\i < 3} {
            \node[squareinactive, minimum width=0.3cm] (ht\i) at (0,0.5*\i) {};
        }{
            \node[squareactive, minimum width=0.3cm] (ht\i) at (0,0.5*\i) {};
        }
    }
    \end{scope}
    
    \begin{scope}[shift={($(ht.east)+(2.5cm,-1cm)$)}, local bounding box=Wxh]
    \foreach \i in {0,...,4} {
        \ifthenelse{\i < 3} {
            \node[squareinactive, minimum width=1.0cm] (Wxh\i) at (0,0.5*\i) {};
        } {
            \node[squareactive, minimum width=1.0cm] (Wxh\i) at (0,0.5*\i) {};
        }
    }
    \end{scope}
    
    \begin{scope}[shift={($(Wxh.east)+(0.7cm,-1cm)$)}, local bounding box=xt]
    \node[squareempty, minimum width=0.3cm] (xt0) at (0,0) {};
    \node[squareempty, minimum width=0.3cm] (xt1) at (0,1.75cm) {};
    \node[squareactive, minimum width=0.3cm, minimum height=1.0cm] (xt) at (0,1.0cm) {};
    \end{scope}
    
    \begin{scope}[shift={($(xt.east)+(2cm,-1cm)$)}, local bounding box=Whh]
    \node[square, minimum size=2.5cm] (rett) at (1,1) {};
    \foreach \i in {0,...,4} {
        \foreach \j in {0,...,4} {
            \pgfmathtruncatemacro{\invj}{4 - \j};
            \ifthenelse{\i < \invj} {
            }{
                \ifthenelse{\invj < 2} {
                    \node[squareactive] (Whh\i\j) at (0.5*\i,0.5*\j) {};
                } {
                    \node[squareinactive] (Whh\i\j) at (0.5*\i,0.5*\j) {};
                }
            }
        }
    }
    \node[textnode] (Whhzero) at (0.25,0.25) {\LARGE\( 0 \)};
    \end{scope}
    
    \begin{scope}[shift={($(Whh.east)+(0.7cm,-1cm)$)}, local bounding box=hprev]
    \foreach \i in {0,...,4} {
        \node[squareactive, minimum width=0.3cm] (hprev\i) at (0,0.5*\i) {};
    }
    \end{scope}

    
    \node[textnode] (textht) at ($(ht.south)+(0,-0.7cm)$) {\( \vm^t \)};
    \node[textnode] (textsigma) at ($(Wxh.south west)+(-0.55cm,-0.7cm)$) {\( \sigma \) \Large \( ( \)};
    \node[textnode] (textequals) at ($(textht.east)!0.5!(textsigma.west)+(0,-0.1cm)$) {\( = \)};
    \node[textnode] (textWxh) at ($(Wxh.south)+(0,-0.7cm)$) {\( \mW^{hm} \)};
    \node[textnode] (textxt) at ($(xt.south)+(0,-1.46cm)$) {\( \vh^{t} \)};
    \node[textnode] (textWhh) at ($(Whh.south)+(0,-0.7cm)$) {\( \mW^{mm} \)};
    \node[textnode] (textplus) at ($(textxt.east)!0.5!(textWhh.west)+(0,-0.05cm)$) {\( + \)};
    \node[textnode] (texthprev) at ($(hprev.south)+(0,-0.7cm)$) {\( \vm^{t-1} \)};
    \node[textnode] (textClose) at ($(hprev.south east)+(0.4,-0.7cm)$) {\Large\( ) \)};
    
    \foreach \i in {0,...,4} {
        \pgfmathtruncatemacro{\invi}{5 - \i};
        \node[textnode] (texthtm\i) at ($(ht.south east)+(0.35,0.25+0.5*\i)$) {\scriptsize\( h_{\invi} \)};
    }
    
    
    \node[textnode] (textht1) at ($(ht.west)+(-0.4,0)$) {\footnotesize\( g N_m \)};
    \node[textnode] (textWh1) at ($(Whh.west)+(-0.4,0)$) {\footnotesize\( g N_m \)};
    \node[textnode] (textWh2) at ($(Whh.north)+(0,0.25)$) {\footnotesize\( g N_m \)};
    \node[textnode] (textxprev1) at ($(hprev.east)+(+0.45,0)$) {\footnotesize\( g N_m \)};
    \node[textnode] (textWx1) at ($(Wxh.west)+(-0.4,0)$) {\footnotesize\( g N_m \)};
    \node[textnode] (textWx2) at ($(Wxh.north)+(0,0.25)$) {\footnotesize\( N_h \)};
    \node[textnode] (textxt1) at ($(xt.east)+(+0.35,0)$) {\footnotesize\( N_h \)};
    
    \end{tikzpicture}
    }
        
    \caption{Representation of the memory update with block matrices showing the size for \( g = 5 \), assuming that only the first two modules are active at time \( t \). Darker blocks represent the active weights.}\label{fig:cwrnn-blocks}
\end{figure}

%% file: img/unrolled_rnn.tex
\definecolor{blue_light}{RGB}{218,232,252}
\definecolor{blue_dark}{RGB}{108,142,191}
\definecolor{green_light}{RGB}{213,232,212}
\definecolor{green_dark}{RGB}{130,179,102}
\definecolor{yellow_light}{RGB}{255,242,204}
\definecolor{yellow_dark}{RGB}{214,182,86}

\begin{tikzpicture}
    \tikzstyle{sty_memory} = [draw=blue_dark,fill=blue_light,circle,outer sep=0,inner sep=1,minimum size=25]
    \tikzstyle{sty_functional} = [draw=green_dark,fill=green_light,circle,outer sep=0,inner sep=1,minimum size=25]
    \tikzstyle{sty_out} = [draw=yellow_dark,fill=yellow_light,circle,outer sep=0,inner sep=1,minimum size=25]
    \tikzstyle{myedgestyle} = [-latex]
    \tikzstyle{sty_module} = [rounded corners,fill opacity=0.4]

    \draw [draw=blue_dark,fill=blue_light,sty_module] (0.25,-2) rectangle (4.75,-0.25);
    \draw [draw=green_dark,fill=green_light,sty_module] (-2,-2) rectangle (0.0,1.75);
    
    \node[] (f_text) at (-1, -1.75) {$Functional$};
    \node[] (f_text) at (2.5, -1.75) {$Memory\ Tape$};
    \node[] (hdot) at (1.6, 1.25) {$\hdots$};
    \node[] (hdot) at (0.35, 0.1) {$\hdots$};
    
    \node[sty_functional] (x) at (-1, -1) {$\vx^t$};
    \node[sty_functional] (h) at (-1, 1) {$\vh^t$};
    
    \node[sty_memory] (hp) at (1, -1) {$\vh^{t-1}$};
    \node[] (hdot) at (2.5, -1) {$\hdots$};
    \node[sty_memory] (hpk) at (4, -1) {$\vh^{t-k}$};
    \node[sty_out] (y) at (1, 3) {$\vy^t$};

    \draw [myedgestyle] (x) edge (h);
    \draw [myedgestyle] (hp) edge (h);
    \draw [myedgestyle] (hpk) edge (h);

    \draw [myedgestyle] (h) edge (y);
    \draw [myedgestyle] (hp) edge (y);
    \draw [myedgestyle] (hpk) edge (y);
\end{tikzpicture}

%% file: img/lmn_train.tex
\definecolor{blue_light}{RGB}{218,232,252}
\definecolor{blue_dark}{RGB}{108,142,191}
\definecolor{green_light}{RGB}{213,232,212}
\definecolor{green_dark}{RGB}{130,179,102}
\definecolor{yellow_light}{RGB}{255,242,204}
\definecolor{yellow_dark}{RGB}{214,182,86}

\begin{tikzpicture}
    \tikzstyle{sty_memory} = [draw=blue_dark,fill=blue_light,circle,outer sep=0,inner sep=1,minimum size=25]
    \tikzstyle{sty_functional} = [draw=green_dark,fill=green_light,circle,outer sep=0,inner sep=1,minimum size=25]
    \tikzstyle{sty_out} = [draw=yellow_dark,fill=yellow_light,circle,outer sep=0,inner sep=1,minimum size=25]
    \tikzstyle{myedgestyle} = [-latex]
    \tikzstyle{sty_module} = [rounded corners,fill opacity=0.4]

    \node[] (f_text) at (1.5, -0.5) {Decoded Memory Tape};
    \node[] (f_text) at (3.8, 3.5) {Decoded Memory Tape};
    
    \node[sty_functional] (x) at (-1, -3) {$\vx^t$};
    \node[sty_functional] (h) at (-1, 1) {$\vh^t$};
    \node[sty_memory] (m) at (4, 1) {$\vm^t$};
    \node[sty_memory] (mp) at (4, -3) {$\vm^{t-1}$};
    \node[sty_out] (y) at (4, 5) {$\vy^t$};
    
    \tikzstyle{tmtape}=[draw,minimum size=0.7cm]
    \begin{scope}[start chain=1 going right,node distance=-0.15mm]
        \node [on chain=1,tmtape] (tape1) at (0, -1) {$\tilde{\vh}^{t - 1}$};
        \node [on chain=1,tmtape] (tape2) {$\tilde{\vh}^{t-2}$};
        \node [on chain=1,tmtape] (tape3) {$\ldots$};
        \node [on chain=1,tmtape] (tape4) {$\tilde{\vh}^{t-k}$};
    \end{scope}
    
    \draw [myedgestyle] (x) edge node[left] {$\mW^{xh}$} (h);
    \draw [myedgestyle] (h) edge node[above] {$\mA$} (m);
    \draw [myedgestyle] (mp) edge node[right] {$\mB$} (m);
    
    \draw [myedgestyle] (mp) edge node[left] {$\mU_k$} (tape3);
    \draw [myedgestyle] (tape3) edge node[right] {$\begin{bmatrix} \mW^{hh}_1 & \hdots & \mW^{hh}_k \end{bmatrix}$} (h);

    \begin{scope}[start chain=2 going right,node distance=-0.15mm]
        \node [on chain=2,tmtape] (2tape1) at (2.3, 3) {$\tilde{\vh}^{t}$};
        \node [on chain=2,tmtape] (2tape2) {$\tilde{\vh}^{t-1}$};
        \node [on chain=2,tmtape] (2tape3) {$\ldots$};
        \node [on chain=2,tmtape] (2tape4) {$\tilde{\vh}^{t-k}$};
    \end{scope}

    \draw [myedgestyle] (m) edge node[left] {$\mU_{k+1}$} (2tape3);
    \draw [myedgestyle] (2tape3) edge node[left] {$\begin{bmatrix} \mW^{hy}_0 & \hdots & \mW^{hy}_k \end{bmatrix}$} (y);
    
    \draw[very thin] (4.5, 1) -- (4.9, 1) -- (4.9, -3);
    \draw[very thin, ->] (4.9, -3) -- (4.5, -3);
    \fill [black] (4.8,-0.5) rectangle (5.,-0.3);

\end{tikzpicture}

%% file: pcode/pseudo_train_lmn.tex
\begin{algorithm}
    \caption{LMN training}\label{alg:lmn_train}
    \begin{algorithmic}[1]
        \Procedure{LMNTrain}{$Data$, $N_h$, $N_m$, $k$}
            \State \(urnn \gets \textit{make-urnn} (N_h, k)\)
            \State \(urnn.fit(Data)\)
            \State \(lmn \gets \textit{init-from-urnn}(urnn, Data, N_m) \)
            \State \(lmn.fit(Data)\)
            \State \Return \(lmn\)
        \EndProcedure
        
        \Procedure{init-from-urnn}{$urnn$, $Data$, $N_m$}
            \State \(\mH \gets []\)
            \For{\(seq \in Data\)} 
                \State \(el \gets urnn(seq)\)
                \State \(\mH.append(el)\)
            \EndFor
            \State \( laes \gets \textit{build-laes}(N_m) \)
            \State \( laes.fit(\mH) \)
            \State \( lmn \gets \textit{build-lmn}(N_h, N_m) \)
            \State \( lmn.\mW^{xh} \gets urnn.\mW^{xh} \)
            \State \( lmn.\mW^{hm} \gets laes.\mA \)
            \State \( lmn.\mW^{mm} \gets laes.\mB \)
            \State \( lmn.\mW^{mh} \gets \begin{bmatrix} urnn.\mW^{hh}_1 & \hdots & urnn.\mW^{hh}_k \end{bmatrix} laes.\mU_k \)
            \State \( lmn.\mW^{mo} \gets \begin{bmatrix} urnn.\mW^{hy}_0 & \hdots & urnn.\mW^{hy}_k \end{bmatrix} laes.\mU_{k+1} \)       
            \State \Return \(lmn\)
            \EndProcedure
    \end{algorithmic}
\end{algorithm}

%% file: img/ms_lmn_init.tex
\definecolor{blue_light}{RGB}{218,232,252}
\definecolor{blue_dark}{RGB}{108,142,191}
\definecolor{green_light}{RGB}{213,232,212}
\definecolor{green_dark}{RGB}{130,179,102}
\definecolor{yellow_light}{RGB}{255,242,204}
\definecolor{yellow_dark}{RGB}{214,182,86}
\definecolor{brick_red}{rgb}{0.8, 0.0, 0.0}

\begin{tikzpicture}
    \tikzstyle{sty_memory} = [draw=blue_dark,fill=blue_light,circle,outer sep=0,inner sep=1,minimum size=25]
    \tikzstyle{sty_functional} = [draw=green_dark,fill=green_light,circle,outer sep=0,inner sep=1,minimum size=25]
    \tikzstyle{sty_out} = [draw=yellow_dark,fill=yellow_light,circle,outer sep=0,inner sep=1,minimum size=25]
    \tikzstyle{myedgestyle} = [-latex]
    \tikzstyle{sty_newedge} = [line width=1pt,draw=brick_red,-latex]
    \tikzstyle{sty_module} = [rounded corners,fill opacity=0.4]

    \draw [draw=black,sty_module] (-2,-2.25) rectangle (2,2);
    \node[] (f_text) at (0, -2) {Trained Model};
    \draw [draw=black,sty_module] (2,-2.25) rectangle (4,2);
    \node[] (f_text) at (3, -2) {New Module};

    \draw [draw=green_dark,fill=green_light,sty_module] (-1.75,-1.75) rectangle (-0.25,1.75);   
    \node[sty_functional] (x) at (-1, -1) {$\vx^t$};
    \node[sty_functional] (h) at (-1, 1) {$\vh^t$};
    \draw [myedgestyle] (x) edge (h);
    \node[sty_out] (y) at (1, 3) {$\vy^t$};

    \draw [draw=blue_dark,fill=blue_light,sty_module] (0.25,-1.75) rectangle (1.75,1.75);
    \node[sty_memory] (m) at (1, 1) {$\vm_1^t$};
    \node[sty_memory] (mp) at (1, -1) {$\vm_1^{t-1}$};
    \draw [myedgestyle, dashed] (h) edge (m);
    \draw [myedgestyle, dashed] (mp) edge (m);
    \draw [myedgestyle] (mp) edge (h);
    \draw [myedgestyle] (m) edge (y);

    \draw [dashed, draw=blue_dark,fill=blue_light,sty_module] (2.25,-1.75) rectangle (3.75,1.75);
    \node[dashed, sty_memory] (m2) at (3, 1) {$\vm_2^t$};
    \node[dashed, sty_memory] (mp2) at (3, -1) {$\vm_2^{t-1}$};
    \draw [sty_newedge, bend left, dashed] (h) edge node[above=1cm, left=.7cm of mp, text=brick_red] {$\mA_g$} (m2);
    \draw [sty_newedge, dashed] (mp2) edge node[right, text=brick_red] {$\mB_g$} (m2);
    \draw [sty_newedge, dashed] (mp2) edge (h);
    \draw [sty_newedge] (m2) edge node[above, right=0.2cm of y, text=brick_red] {$\mW^{out}_g$} (y);
    \draw [sty_newedge, dashed] (mp2) edge (m);
    
    \draw[very thin] (1.5, 1) -- (1.9, 1) -- (1.9, -1);
    \draw[very thin, ->] (1.9, -1) -- (1.5, -1);
    \fill [black] (1.8,-0.5) rectangle (2.,-0.3);
    
    \draw[very thin] (3.5, 1) -- (3.9, 1) -- (3.9, -1);
    \draw[very thin, ->] (3.9, -1) -- (3.5, -1);
    \fill [black] (3.8,-0.5) rectangle (4.,-0.3);
    
\end{tikzpicture}

%% file: pcode/pseudo_train_ms_lmn.tex
\begin{algorithm}
    \caption{MS-LMN training}\label{alg:ms_lmn_train}
    \begin{algorithmic}[1]
        \Procedure{MS-LMNTrain}{$Data$, $N_h$, $N_m$, $g$}
            \State \( \textit{ms-lmn} \gets \textit{init-single-module}(N_h, N_m) \)
            \State \( \textit{ms-lmn.fit}(Data) \)
            \For {\( i \in \{2, \hdots, g\} \)}
                \State \( \textit{ms-lmn.add-module}(Data, i) \)
                \State \( \textit{ms-lmn.fit}(Data) \)
            \EndFor
            \State \Return \( \textit{ms-lmn} \) 
        \EndProcedure
        
        \Procedure{add-module}{$self$, $Data$, $i$}
            \State \( \mH_g = [] \)
            \For {\(seq \in Data\)}
                \State \( \vh_{seq}, \vm_{seq} \gets self(seq) \)
                \State \( \vh_{seq}.subsample(2^g)\)
                \State \( \mH_g.append(\vh_{seq}) \)
            \EndFor
            \State \( laes \gets \textit{build-laes}(N_m) \)
            \State \( laes.fit(\mH_g) \)
            \State \( self.\mW^{h m_{g+1}} \gets laes.\mA \)
            \State \( self.\mW^{m_{g+1} m_{g+1}} \gets laes.B \)
            \State \( self.\mW^{m_{g+1} o} \gets \mW_{g+1} \)
        \EndProcedure
    \end{algorithmic}
\end{algorithm}

%% file: img/gen_seq.tex
\begin{tikzpicture}

\node[label={RNN (NMSE=$79.5$)},draw=none,fill=none] (rnn) at (0,-1.5){\includegraphics[width=.5\textwidth]{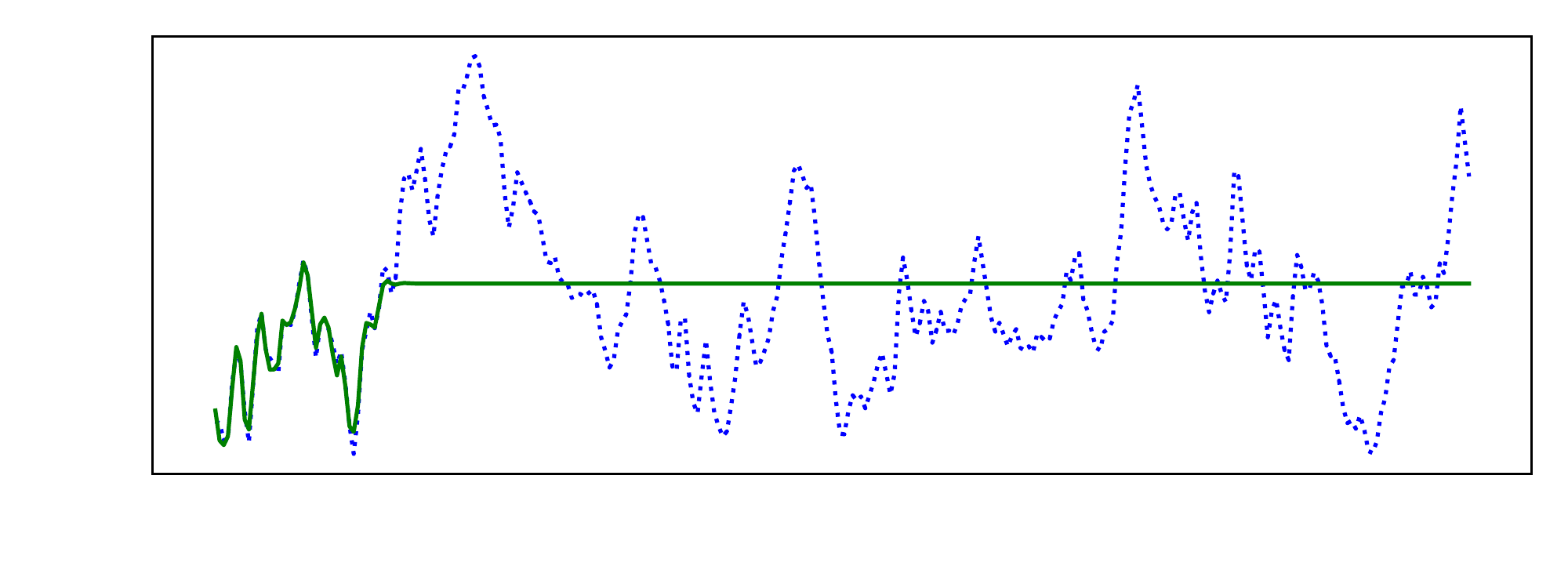}};
\node[label={LMN (NMSE=$38.4$)},draw=none,fill=none] (lmn) at (0, -4.5) {\includegraphics[width=.5\textwidth]{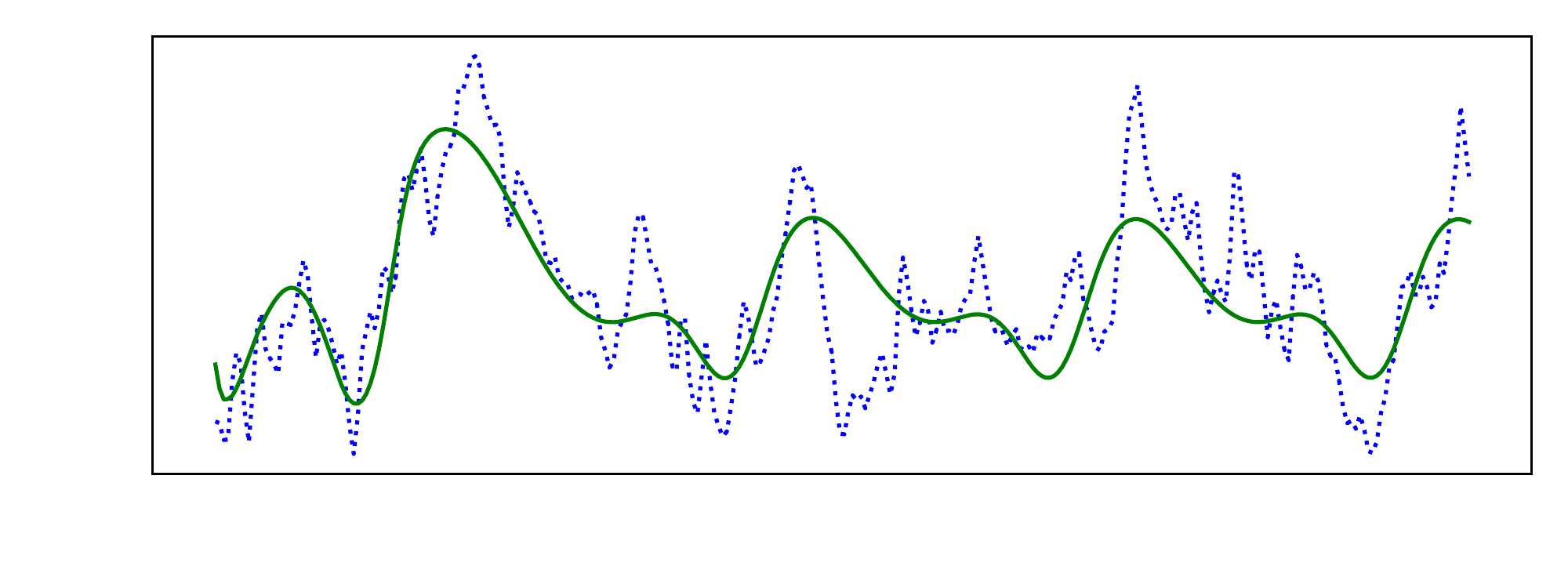}};

\node[label={CW-RNN (NMSE=$12.5$)},draw=none,fill=none] (cwrnn) at (6, -1.5) {\includegraphics[width=.5\textwidth]{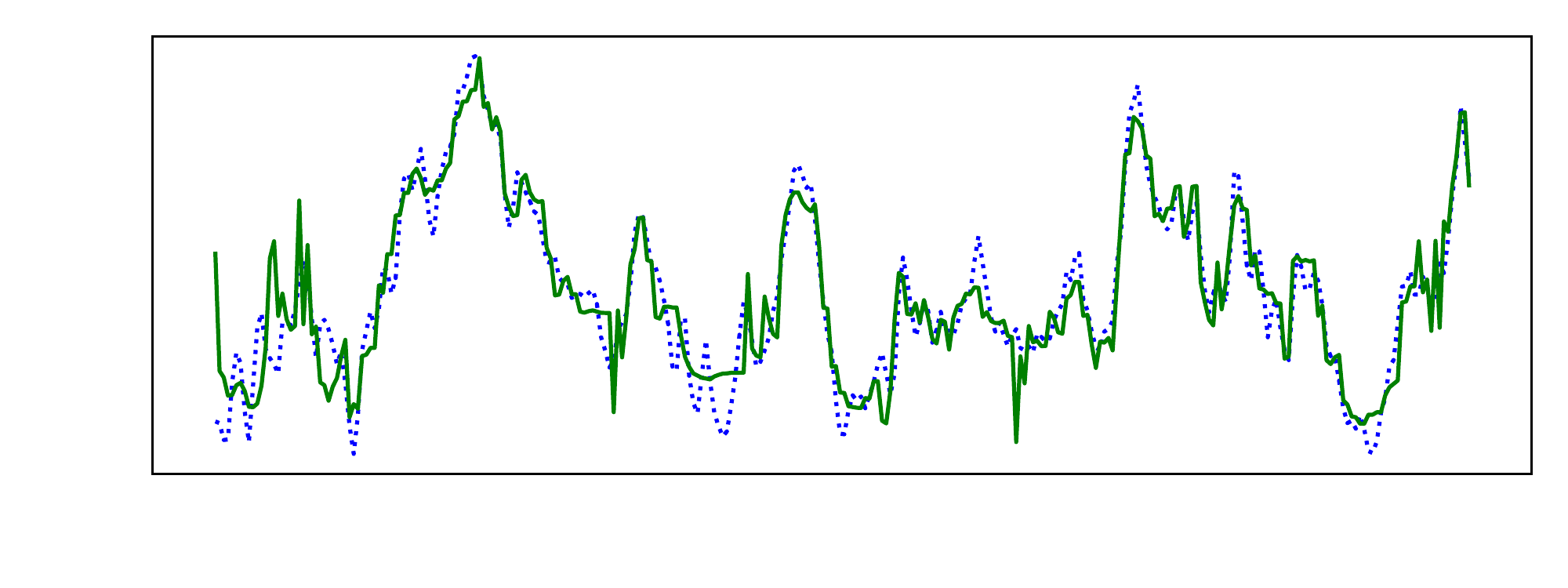}};
\node[label={MS-LMN (NMSE=$0.116$)},draw=none,fill=none] (cw-lmn) at (6, -4.5) {\includegraphics[width=.5\textwidth]{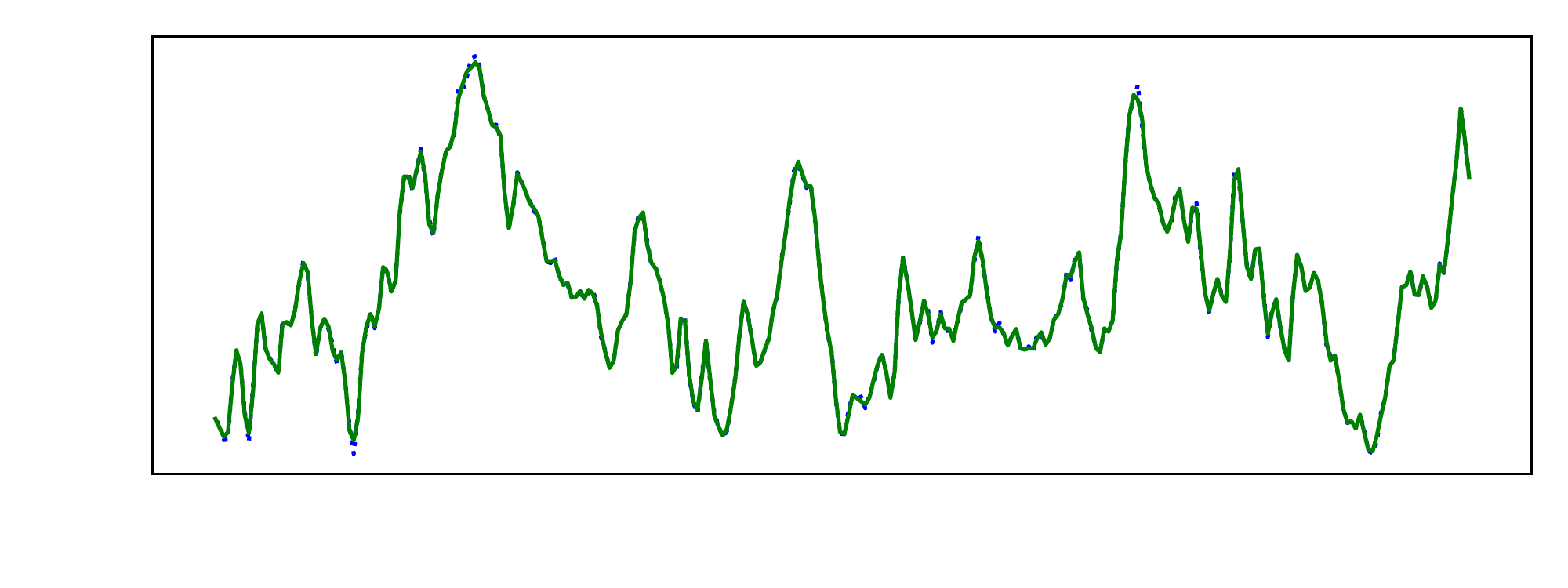}};

\end{tikzpicture}

%% file: appendix.tex
\section{MS-LMN equations}
Similarly to Eq. \ref{eq:mslmn_m_block1}-\ref{eq:mslmn_m_block2}, we can modify the MS-LMN computation of $\vh^t$ and $\vy^t$ shown in Equations \ref{eq:ms_lmn_h} and \ref{eq:ms_lmn_out} to allow an efficient parallel implementation. The parameters of the modules are combined together as follows:
\begin{align*}
    \mW^{mh} &= \begin{bmatrix} 
        \mW^{m_1 h} & \hdots & \mW^{m_g h}
    \end{bmatrix} \\
    \mW^{my} &= \begin{bmatrix} 
        \mW^{m_1 y} & \hdots & \mW^{m_g y}
    \end{bmatrix}.
\end{align*}
Equations \ref{eq:ms_lmn_h} and \ref{eq:ms_lmn_out} become:
\begin{align*}
    \vh^t &= \sigma(\mW^{xh} \vx^{t} + \mW^{mh} \vm^{t-1}) \\
    \vy^t &= \mW^{m y} \vm^t.
\end{align*}
Notice that, unlike Eq. \ref{eq:mslmn_m_block1}-\ref{eq:mslmn_m_block2}, the subsampling is not required to compute $\vh^t$.

\section{TIMIT Preprocessing}
To help with the reproducibility of the Common Suffix TIMIT task we report for each word the code corresponding to the files in the original datasets which have been used.
Although there may be more samples for some classes, to compare with the experiments from \cite{clockwork_koutnik14} we chose to keep the classes balanced. Therefore, we indicate the file that we used to build our dataset in the following table.
    
\begin{tabular}{ll|ll|ll}
    \toprule
    Word     & Code    & Word   & Code     & Word      & Code \\
    \midrule
    Making & SX430 & Classical & SX52 & Discussions  & SX40 \\
    Walking & SX320 & Critical & SX52 & Regulations & SX41 \\
    Cooking & SX177 & Tradition & SX137 & Accusations & SX191 \\
    Looking & SX229 & Addition & SX45 & Conditions & SX349 \\
    Working & SX4 & Audition & SX194 & Subway & SX246 \\
    Biblical & SX42 & Recognition & SX251 & Leeway & SX230 \\
    Cyclical & SX146 & Competition & SX141 & Freeway & SX233 \\
    Technical & SX135 & Musicians & SX15 & Highway & SX233 \\
     & & & & Hallway & SX106 \\
    \bottomrule
\end{tabular}